\documentclass{article}

\usepackage[nonatbib,preprint]{neurips_2025}

\usepackage{graphics}
\usepackage[utf8]{inputenc} 
\usepackage[T1]{fontenc}    
\usepackage{hyperref}       
\usepackage{url}            
\usepackage{booktabs}       
\usepackage{amsmath,amsfonts,amsthm}
\usepackage{nicefrac}       
\usepackage{microtype}      
\usepackage{xcolor}         
\usepackage{algorithm}
\usepackage{algpseudocode}
\usepackage{pifont}

\usepackage{graphicx}
\usepackage{subcaption}  

\usepackage{biblatex}
\usepackage{bbm}

\newcommand{\nulll}{\textsf{null}}
\newtheorem{theorem}{Theorem}[section]
\newtheorem{assumption}[theorem]{Assumption}
\newtheorem{lemma}[theorem]{Lemma}
\newtheorem{proposition}[theorem]{Proposition}
\newtheorem{cor}[theorem]{Corollary}

\title{Episodic Contextual Bandits with Knapsacks under Conversion Models}

\author{%
  Wang Chi Cheung\\
  Industrial Systems Engineering \& Management\\
  National University of Singapore\\
  Engineering Drive 2 Block E1A 06-25 Singapore 117576\\
  \texttt{isecwc@nus.edu.sg} \\
  \And
  Zitian Li\\
  Industrial Systems Engineering \& Management\\
  National University of Singapore\\
  Engineering Drive 2 Block E1A 06-25 Singapore 117576\\
  \texttt{lizitian@u.nus.edu} \\
}

\begin{document}

\maketitle

\begin{abstract}
  We study an online setting, where a decision maker (DM) interacts with contextual bandit-with-knapsack (BwK) instances in repeated episodes. These episodes start with different resource amounts, and the contexts' probability distributions are non-stationary in an episode. All episodes share the same latent conversion model, which governs the random outcome contingent upon a request's context and an allocation decision. Our model captures applications such as dynamic pricing on perishable resources with episodic replenishment, and first price auctions in repeated episodes with different starting budgets. We design an online algorithm that achieves a regret sub-linear in $T$, the number of episodes, assuming access to a \emph{confidence bound oracle} that achieves an $o(T)$-regret. Such an oracle is readily available from existing contextual bandit literature. We overcome the technical challenge with arbitrarily many possible contexts, which leads to a reinforcement learning problem with an unbounded state space. Our framework provides improved regret bounds in certain settings when the DM is provided with unlabeled feature data, which is novel to the contextual BwK literature. 
\end{abstract}

\section{Introduction}
Contextual bandits with knapsacks (Contextual BwKs) is a fundamental model for contextual online resource allocation under model uncertainty. Its applications include personalized dynamic pricing \cite{BanK21,LiZ23,WangTL24}, personalized loan discounts \cite{LiS22}, dynamic procurement \cite{BadanidiyuruKS12,BadanidiyuruKS13}, and first price auctions \cite{WangYDK23,CastiglioniCK24} (see Appendix \ref{app:application} for details). In existing research, contextual BwKs are only studied in the single episode setting. In an episode, the decision maker (DM) begins with certain amounts of resources, and then makes allocation decisions contingent upon each arriving request's context. While the single episode setting is of great importance, in practice, contextual online resource allocation is performed over multiple episodes, where the resources are restocked after an episode ends. For example, in personalized dynamic pricing of a product, the product units are restocked after an episode of the current selling process ends, in anticipation of the next episode. In first price auctions for online ad allocations, an advertiser restores his/her advertising budget after exhausting the current budget. 

When the DM makes allocation decisions and observes their outcomes in multiple episodes, the DM is accumulating data on the underlying conversion model, which maps a request's context and an allocation decision to the corresponding expected outcome. Can the DM converge to optimality, as the number of episodes increases? We shed a positive light on this question. 
We first observe that existing works on the single episode case cannot achieve the desired convergence, see forthcoming Section \ref{sec:main}. Different from existing works on contextual BwKs, we directly analyze the Bellman equation associated with the contextual BwK problem in an episode, which is a reinforcement learning (RL) problem with an intractable state space. Specifying the transition probability requires knowing the probability distributions of the contexts, which requires a sample size linear in the number of contexts. Despite the difficulty, we achieve a regret bound independent of the number of contexts, by a careful consideration on how estimation errors propagate as we backtrack the value-to-go functions. Instead of assuming a specific contextual reward and resource consumption model, our algorithm framework accommodates a variety of contextual multi-armed bandit models, such as contextual linear bandits \cite{AbbasiPS11}, contextual generalized linear bandits \cite{FilippiCGS10}, contextual Gaussian process bandits \cite{KrauseO11}, kernelized bandits \cite{ChowdhuryG17}, neural bandits \cite{pmlr-v119-zhou20a}, via the notion of \emph{confidence bound oracles}, see Section \ref{sec:main}. Our analytical framework demonstrates the benefits of having sufficiently many unlabeled feature data, which is novel in contextual BwK settings.

\textbf{Related Works. }Our contributions are related to the research works on contextual BwKs, online reinforcement learning (RL) on large state spaces, and contextual online RL.

The BwK problem is first introduced by \cite{BadanidiyuruKS13}. Its generalization to concave rewards and convex constraints is studied in \cite{AgrawalD14}, and its adversarial variant is studied in \cite{ImmorlicaSSS22}. The contextual multi-armed bandit generalization of BwKs, dubbed contextual BwKs, is first studied in \cite{BadanidiyuruKS18}.  Its generalization to concave rewards and convex constraints is studied by \cite{AgrawalDL16}. \cite{AgrawalD16} provides an efficient algorithm for linear contextual BwKs. \cite{LiS22} proposes the contextual BwK model for a conversion model, which is incorporated in our model formulation. \cite{LiS22} models the conversion model by the contextual logistic bandit model. In our setting, we show that a variety of confidence-bound based research works on contextual multi-armed bandit models, such as the contextual linear bandit models \cite{AbbasiPS11}, the contextual generalized linear  bandit models \cite{FauryACF2020}, its specialization the contextual logistic bandit models \cite{FauryACF2020}, the contextual Gaussian process bandit models \cite{KrauseO11}, kernelized bandit model \cite{ChowdhuryG17}, neural bandit model \cite{pmlr-v119-zhou20a}, can be used to model the conversion model and achieve non-trivial regret bounds, by constructing the corresponding \emph{confidence bound oracles} based on those works. A recent stream of research works \cite{HanZWXZ23,SlivkinsZSF24,GuoL24,GuoL25} studies the contextual BwKs assuming access to the online regression oracle \cite{FosterR20}, which is different and in general incomparable to our oracle assumption. Contextual BwKs have also been studied in dynamic pricing settings with inventory constraint settings \cite{LiZ23,WangTL24}. We provide a detailed comparison with these works on contextual BwKs in Section \ref{sec:main}. Lastly, \cite{ChenAYPWD24} considers the application of resolving heuristics on an online contextual decision making model with knapsack constraints. While they achieve a logarithmic regret compared with the optimum under mild assumptions on the corresponding fluid relaxation (see Appendix \ref{app:supp_CBwK}), they require either the full feedback setting, or a partial feedback setting that is strictly more informative than the bandit feedback setting. 

Our online contextual resource allocation problem can be formulated as an online RL problem with large state spaces. A cornucopia of research works have been done on function approximation on the value-to-go functions by (generalized) linear functions \cite{YangW19,WangWDK21,ChowdhuryGM21,LiLKSWY22}, and general functions via the structural conditions of Eluder Dimensions \cite{RussoVR13,OsbandRV14}, Bellman Eluder dimension \cite{JinLM21}, Bilinear classes \cite{DuKLLMSW21}, Decision-Estimation Coefficient \cite{FosterKQR23}, and Admissible Bellman Characterization \cite{ChenLYGJ23}. There have been substantial works on function approximation on the transition probability kernels instead of the value-to-go functions. \cite{JinYZJ20} study the case of linear function approximation, while \cite{hwang2023model,ChoHLO24,LiZZZ24} study the case of multinomial logit function approximation. While our online RL problem also suffers from the intractability with large state space, our underlying value-to-go functions and transition probability kernels (see Section \ref{sec:bellman}) do not match the approximations mentioned above. A crucial challenge lies in the unknown probability distributions in the contexts. These probability distributions do not enjoy the structural conditions in the above works, which motivate us to take a different route for overcoming the intractability.

Lastly, our work is also related to contextual online RL, studied in \cite{FosterKQR23,LevyM23,DengCZL24,QianHSC24,FosterHQR24}. In these models, there is only one context per episode. The context is generated before the episode begins, and the rewards and transition probability kernels in each time step during the epsiode are all dependent on that context. This is different from our model, which has $H$ contexts for the $H$ time steps during an episode, and each the context at time step $h$ only appears after time step $h-1$ but before time step $h$. In addition, by harnessing the problem structure with resource allocation, we achieve a better dependence on $H$ then those works, though their models and ours are not comparable in general.

\textbf{Notation.} We denote $\mathbb{R}, \mathbb{Z}$ as the sets of real numbers and integers respectively, and denote $\mathbb{R}_{>0}, \mathbb{Z}_{>0} $ as the sets of positive real numbers and positive integers respectively. For $n\in \mathbb{Z}_{>0}$, denote $[n] = \{1, \ldots, n\}$. For any $q\in [0, 1]$, denote $\text{Bern}(q)$ as the Bernoulli distribution with mean $q$. For an event ${\cal E}$, denote $\mathbf{1}({\cal E})$ as its indicator random variable, where $\mathbf{1}({\cal E})=1$ if event ${\cal E}$ happens, and $\mathbf{1}({\cal E})=0$ otherwise. For $a, b\in \mathbb{R}$, denote $a\wedge b = \min\{a, b\}$. Denote the size of a set ${\cal A}$ as $|{\cal A}|$.

\section{Problem Formulation }
\subsection{Model}\label{sec:model}
We study an online optimization problem with $T$ episodes. Each episode consists of a contextual BwK problem with $H\in \mathbb{Z}_{>0}$ time steps. Our model captures applications such as personalized dynamic pricing \cite{WangTL24}, contextual dynamic procurement \cite{BadanidiyuruKS13}, and repeated first-price auctions with budget constraints \cite{BalseiroGMMS19,WangYDK23,CastiglioniCK24,HanWz25}, in multiple episodes. 
We provide the formulation of our model, then highlight the mapping from our model to the applications in Table \ref{tab:application}, with the details provided in the Appendix.

\textbf{Parameters.} 
For each $h\in [H], t\in [T]$, denote the $h$-th time step in the $t$-th episode as $(h, t)$. Over the horizon of $HT$ time steps, these time steps are ordered lexicographically as $(1, 1), \ldots, (H, 1), (1, 2), \ldots (2, T), (3, 1), \ldots, (H, T)$. Episode $t$ consists of time steps $(1, t), \ldots, (H, t)$. There is one request that arrives at each time step. Each request is associated with a context $\theta$, encoding the characteristics of the request. The set of all contexts is denoted as $\Theta$, which can be infinite. Denote ${\cal A}$ as the finite set of allocation decisions. 

Each $(\theta, a)\in \Theta\times {\cal A}$ is associated with three deterministic quantities: (i) $\rho(\theta, a)\in [0, 1]$, the conversion probability (ii) $r(\theta, a)\in (-\infty, r_\text{max}]$, the amount of reward (iii) $d(\theta, a)\in \{0\}\cup [L]$, the amount of resource consumption. There is a null context $\theta_{\nulll}\in \Theta$ that represents the case of no request arrival, where $r(\theta_{\nulll}, a) = d(\theta_{\nulll}, a)=\rho(\theta_{\nulll}, a)=0$ for all $a\in {\cal A}$. There is a null action $a_{\nulll}\in {\cal A}$ that represents the choice of no allocation, where $r(\theta, a_{\nulll}) = d(\theta, a_{\nulll})=\rho(\theta, a_{\nulll})=0$ for all $\theta\in \Theta$. For each $b\in \{0\}\cup [LH], \theta\in \Theta$, denote $${\cal A}(b, \theta) = \{a\in {\cal A} : r(\theta, a)\geq 0\text{, and }d(\theta , a) \leq b\}.$$ 
Note that $a_{\nulll} \in {\cal A}(b, \theta)$ for any $b, \theta$, and $r(\theta, a)\in [0, r_\text{max}]$ for any $a\in {\cal A}(b, \theta)$.


\textbf{Dynamics.} At the start of episode $t$, the DM is endowed with $B_t\in [LH]$ units of the resource. The DM only observes the value of $B_t$ after the end of episode $t-1$, but before the start of episode $t$. 
Denote $b_{h, t}$ as the amounts of the remaining resource at the start of the time step $(h, t)$ (observe that $b_{1, t} = B_t$). 
At the time step $(h, t)$, 
firstly a request with context $\theta_{h, t}\in \Theta$ appears. The context $\theta_{h, t}$ is distributed according to the latent probability distribution $\Lambda_h$. The contexts $(\theta_{h,t})^H_{h=1} \sim \prod^H_{h=1}\Lambda_h$ are independent but not necessarily identically distributed, since $\Lambda_1, \ldots, \Lambda_H$ vary in general. 
After that, the DM selects an allocation decision $A_{h, t} \in {\cal A}(b_{h, t}, \theta_{h, t})$,  based on $\theta_{h, t}$, $b_{h, t}$ and the observations obtained prior to time step $(h, t)$. The restricted set ${\cal A}(b_{h, t}, \theta_{h, t})$ ensures that the DM consumes at most $B_t$ resource units in episode $t$, and does not violate the resource constraint.

After $A_{h, t}$ is selected, the nature samples $Y_{h,t }\sim \text{Bern}(\rho(\theta_{h,t}, A_{h, t}))$, and $Y_{h,t }$ is revealed to the DM. The realization of $Y_{h, t} = 1$ means that the request accepts the allocation, but $Y_{h, t} = 0$ means that the request rejects. That is, $\rho(\theta_{h,t}, A_{h, t})$ is the conversion probability of the request at $(h,t)$. The DM earns a reward of $R_{h, t} = r(\theta_{h, t}, A_{h, t}) \cdot Y_{h,t}$, but consumes $D_{h, t} = d(\theta_{h, t}, A_{h, t}) \cdot Y_{h,t}$ units of the resource. Lastly, if $h<H$, then the DM ends time step $(h, t)$, proceeds to time step $(h+1, t)$ within episode $t$, and updates $b_{h+1, t} = b_{h, t} - D_{h, t} \geq 0$. Otherwise, we have $h = H$, and episode $t$ has ended. The remaining $b_{H+1, t}$ resource units are either scrapped or clawed back to the DM depending on the application (see Appendix \ref{app:application}), but they are not carried over to the next episode. Instead, in the next episode $t+1$, the DM starts at time step $(1, t+1)$ with a new budget of $b_{1, t+1} = B_{t+1}$ units of the resource. Lastly, we assume $B_t\in [LH]$ without loss of generality, since $LH$ is the maximum amount of the resource the DM can consume in an episode.
 
\textbf{Knowledge on Model.} The DM only has partial knowledge about the underlying model. On the one hand, the DM knows $H$, and the bounding parameters $r_\text{max}, L$. The DM knows $\{r(\theta, a), d(\theta, a)\}_{\theta\in \Theta, a\in {\cal A}}$, and the feasible set ${\cal A}(b, \theta)$ for each $b\in \{0\}\cup [LH], \theta\in \Theta$. On the other hand, the DM does not know the probability distributions $\Lambda_1, \ldots, \Lambda_H$ that generate the contexts. The DM does not know the conversion probability $\rho(\theta, a)$ for any $\theta\in \Theta\setminus\{\theta_{\nulll}\}, a\in {\cal A}\setminus\{a_{\nulll}\}$.The DM knows that $\rho(\theta_\nulll, a) = \rho(\theta, a_\nulll) = 0$ for any $\theta\in \Theta, a\in {\cal A}$. The sequence $B_1, \ldots, B_T\in [LH]$ is set by an oblivious adversary and is arbitrarily changing, and $B_t$ is only revealed at the start of episode $t$.  
Table \ref{tab:application} presents the practical motivation behind the model, particularly in knowing ${\cal A}(b, \theta), r(\theta,a), d(\theta, a)$ but not $\rho(\theta, a)$. 

We make the following assumption on the DM's ability to learn the underlying conversion model $\rho$.
\begin{assumption}\label{ass:conf}
The DM has access to a confidence bound (CB) oracle ${\cal O}$. The oracle ${\cal O}$ inputs a confidence parameter $\delta\in (0,1)$ and a sequence of online data ${\cal D} = \{ (\theta_n, A_n, Y_n) \}^N_{n=1}$, where $(\theta_1, A_1), \ldots, (\theta_N, A_N)\in \Theta\times {\cal A}\setminus \{a_\nulll\}$ are arbitrarily generated, and $Y_n\sim \text{Bern}(\rho(\theta_n, A_n))$. The oracle outputs two functions $\text{UCB}, \text{LCB}: \Theta\times {\cal A}\times(0,1) \rightarrow [0,1]$ such that 
$$
\Pr\left(\rho(\theta, a)\in [\text{LCB}(\theta, a;\delta), \text{UCB}(\theta, a;\delta)] \text{ for all $\theta\in \Theta, a\in {\cal A}$}\right)\geq 1-\delta.
$$
In addition, the output satisfies $\text{UCB}(\theta, a_\nulll;\delta) =  \text{LCB}(\theta, a_\nulll;\delta) = 0$ for all $\theta\in \Theta$.
\end{assumption}
A CB oracle can be directly constructed based on the existing literature on contextual multi-armed bandits, when the conversion conversion probability $\rho$ is modeled by the generalized linear contextual bandit model \cite{FilippiCGS10}, contextual logistic bandit model \cite{FauryACF2020,LiS22}, contextual Gaussian bandit model \cite{KrauseO11}, etc.
We provide details on these oracles in Appendix \ref{app:oracles}. Lastly, we make a mild assumption that, to receive a positive reward, some unit(s) of the resource has to be consumed.
\begin{assumption}\label{ass:nonnull}
For all $a\in {\cal A}\setminus \{a_\nulll\}$, we have $d(\theta, a) > 0$ for any $\theta\in \Theta$.
\end{assumption}

\begin{table}
\centering
\begin{tabular}{|c||c|c|c|}
\hline 
& Pricing & Procurement & First Price Auction \\ 
\hline 
$B_t$ & product inventory & Compensation budget & bidding budget \\ 
\hline
Requests & customers & workers & auction items \\ 
\hline 
$\theta$ & customer's feature & worker's feature & DM's private value \\ 
\hline 
$a$ & offered price & offered price  & DM's bid \\ 
\hline 
${\cal A}(b, \theta)$ & ${\cal A}$ if $b\geq 1$, & $\{a_{\nulll}=0\} \cup[b\wedge L]$  & $\{a\in \{a_\nulll = 0\}\cup [L]: $  \\
& $\{a_{\nulll}\}$ if $b=0$ &  & $a\leq b, a\leq \theta\}$\\
\hline
$\rho(\theta, a)$ & $\Pr(\text{customer buys})$ & $\Pr(\text{worker accepts})$ & $\Pr(\text{DM wins the auction})$ \\ 
\hline 
$r(\theta, a)$ & $a$ & $1$ & $\theta-a$ \\ 
\hline 
$d(\theta, a)$ & $1$ & $a$ & $a$ \\ 
\hline 
\end{tabular} 
\caption{A summary on the three applications. Each entry corresponds to the meaning in the application.}
\label{tab:application}
\end{table}
 

\textbf{Objective.} Denote $\textsf{opt}_t$ as the optimal expected total reward in episode $t$, and $\textsf{opt}_t$ varies with $B_t$. We aim to design an online policy over the $T$ episodes (total $HT$ time steps) whose regret
$$
\text{Reg}(T) = \sum^T_{t=1}\left[\textsf{opt}_t - \sum^H_{h=1} R_{h, t} \right]=o(T),
$$
meaning $\text{Reg}(T)$ grows sub-linearly with $T$, with high probability.  Our aim is different from conventional research works on contextual BwKs, who focus on one episode and aim to achieve $\textsf{opt}_1 - \sum^H_{h=1} R_{h, 1} = o(H)$. By achieving an $o(T)$-regret, the average optimality gap $ (1/T)\sum^T_{t=1} [\textsf{opt}_t -\sum^H_{h=1} R_{h, t}]$ per episode shrinks to 0 as $T$ grows. The shrinkage is not established in existing works on contextual BwKs, as elaborated in the forthcoming Section \ref{sec:main}. Our objective of $o(T)$-regret is in line with the objective in online episodic reinforcement learning, though we face a unique challenge with uncertainty over a large state space, due to the uncertainty on $\Lambda_1, \ldots, \Lambda_H$ over an arbitrarily large context space $\Theta$ and the uncertainty on $\rho$, as explained below.

\subsection{Characterizing $\textsf{opt}_1\ldots, \textsf{opt}_T$ with Bellman Equations}\label{sec:bellman}
For episodes $1, \ldots, T$, their optima $\textsf{opt}_1, \ldots, \textsf{opt}_T$ can be achieved by a policy $\pi^*$, which selects an optimal action $\pi^*_h(b, \theta)$ contingent upon the remaining resource amount $b$ and the request's context $\theta$ at the start of time step $h$. For any $h\in [H], b\in \{0\}\cup [LH], \theta\in \Theta$, the action $\pi^*_h(b, \theta)\in {\cal A}(b, \theta)$ is an optimal solution to the maximization problem on the right hand side of
\begin{align}
V_h(b, \theta) = \max_{a\in {\cal A}(b,\theta)}&\left\{\rho(\theta, a) r(\theta, a) +  \sum_{\bar{\theta}\in \Theta} \rho(\theta, a)\cdot \Pr_{\theta_{h+1}\sim \Lambda_{h+1}}(\theta_{h+1} = \bar{\theta})\cdot V_{h+1}(b - d(\theta, a), \bar{\theta})  \right.\nonumber\\
&\qquad \left. +\sum_{\bar{\theta}\in \Theta}  (1-\rho(\theta,a)) \cdot\Pr_{\theta_{h+1}\sim \Lambda_{h+1}}(\theta_{h+1} = \bar{\theta})\cdot V_{h+1}(b, \bar{\theta})\right\},
\label{eq:bellman}
\end{align}
and we set $V_{H+1}(b,\theta) = 0$ for all $b,\theta$. Observe $\textsf{opt}_t = \sum_{\bar{\theta}\in \Theta} \Pr_{\theta_{1}\sim \Lambda_{1}}(\theta_{1} = \bar{\theta})\cdot V_{1}(B_t, \bar{\theta})$. The set of Bellman equations (\ref{eq:bellman}) highlights the challenge with the large state space $(\{0\}\cup [LH])\times\Theta$. 
The latent transition probability is of the form $\rho(\theta, a)\cdot \Pr_{\theta_{h+1}\sim \Lambda_{h+1}}(\theta_{h+1} = \bar{\theta})$, thus the number of latent parameters is linear in $H\cdot |{\cal A}|\cdot |\Theta|$, which is intractably large. A reason of the intractability is the term  $\Pr_{\theta_{h+1}\sim \Lambda_{h+1}}(\theta_{h+1} = \bar{\theta})$, which is in general unstructured, and typical functional approximation techniques in RL doe not apply. Lastly, the above discussion assumes a finite or countably infinite $\Theta$, but we can handle an uncountably infinite (but measurable) $\Theta$ by replacing $\sum_{\bar{\theta}\in \Theta} \Pr_{\theta_{h+1}\sim \Lambda_{h+1}}(\theta_{h+1} = \bar{\theta})\cdot V_{h+1}(b', \bar{\theta})$ with $\mathbb{E}_{\theta_{h+1}\sim \Lambda_{h+1}}[V_{h+1}(b', \theta_{h+1})]$, which we denote as $U_{h+1}(b')$.

\section{Online Algorithms}

\subsection{Algorithm Design}
We propose the online algorithm \textsf{Mimic-Opt-DP}, shown in Algorithm \ref{alg:online}. We first explain the main steps, then discuss how \textsf{Mimic-Opt-DP} estimates the latent parameters. \textsf{Mimic-Opt-DP} involves two sets of estimates: (i) $\{\hat{U}_{h+1, t}\}_{h\in [H+1]}$: $\hat{U}_{h+1, t}(b)$ serves to estimate $U_{h+1}(b) = \mathbb{E}_{\theta_{h+1}\sim \Lambda_{h+1}}[V_{h+1}(b, \theta_{h+1})]$, the optimal expected reward from time step $h+1$ to $H$, when the DM has $b$ units of remaining resources at the start of time step $h+1$. (ii) $\text{UCB}_t, \text{LCB}_t$, which serves to estimate $\rho$. The estimates $\hat{U}_{h+1, t}$, $\text{UCB}_t, \text{LCB}_t$, constructed using data in episodes $1, \ldots, t-1$, are for choosing the action $A_{h, t}$ at time step $(h, t)$ (Line \ref{alg:online_Aht} in Algorithm \ref{alg:online}). The optimization problem (\ref{eq:alg_online_Aht}) serves to estimate the Bellman equation (\ref{eq:bellman}). To formulate (\ref{eq:alg_online_Aht}), we replace the latent $\rho$ with its optimistic estimates, where we use upper confidence bounds (UCBs) for the reward earned and lower confidence bounds (LCBs) for the resource consumed. The replacement thus follows the principle of Optimism in Face of Uncertainty.

We construct $\{\hat{U}_{h, t}\}_{h\in [H+1]}$ in $\textsf{Optimistic-DP}$ displayed in Algorithm \ref{alg:UCB}. The construction, shown in Lines \ref{alg:UCB_for_start} to \ref{alg:UCB_for_end} in Algorithm \ref{alg:UCB}, involves a backward induction that resembles the Bellman equations (\ref{eq:bellman}). We replace the latent parameters in (\ref{eq:bellman}) with their estimates similarly to (\ref{eq:alg_online_Aht}). In this way, the action selection in (\ref{eq:alg_online_Aht}) in fact mimics the way $\{\hat{U}_{h, t}\}_{h\in [H+1]}$ are constructed. The backward induction in Algorithm \ref{alg:UCB} requires estimating both the latent quantities $\rho$ and $\{\Lambda_h\}_{h\in [H]}$. At the end of episode $t-1$, we estimate $\rho$ and $\{\Lambda_h\}_{h\in [H]}$ using two disjoint datasets ${\cal D}_t$ and ${\cal S}_t$ respectively. Their disjointness facilitates our analysis. We use the data in episodes in ${\cal J}_t\subset [t-1]$ to estimate $\{\Lambda_h\}_{h\in [H]}$, and the data in episodes in $[t-1] \setminus {\cal J}_t$ to estimate $\rho$. Our default choice of $\{{\cal J}_t\}^\infty_{t=1}$ depends on an absolute constant $\alpha\in (0,1)$, which we take $\alpha=1/2$ by default. Let $n_\alpha\in \mathbb{Z}_{>0}$ satisfy $(1/\alpha) - 1 \leq n_\alpha < (2/\alpha)- 1$. We define the default choice
\begin{equation}\label{eq:J}
    {\cal J}_\infty = \{1+ i\cdot n_\alpha: i\in \mathbb{Z}_{\geq 0}\}, \quad {\cal J}_{t} = {\cal J}_\infty\cap [t-1]\quad \text{for each $t\in \mathbb{Z}_{>0}$.}
\end{equation}
We design ${\cal J}_{t+1}$ so that $\alpha t/2 \leq |{\cal J}_{t+1}|\leq \alpha t$, and consecutive indexes in ${\cal J}_t$ are $n_\alpha$ apart. The design ensures that the amounts of data for learning $\{\Lambda_h\}_{h\in [H]}$ and $\rho$ both increase linearly with $t$. 

For estimating $\{\Lambda_h\}_{h\in [H]}$, the DM collates the unlabeled feature data $(\theta_{h, s})^H_{h=1}$ after each episode $s\in {\cal J}_t\subset [t-1]$ to form the dataset ${\cal S}_t$ (Line \ref{alg:collate_unlabel}). In the forthcoming Sections \ref{sec:main}, \ref{sec:highlevel}, we focus on the case when the DM starts with no feature data (i.e. ${\cal S}_1=\emptyset$, hence $|{\cal S}_t| = |{\cal J}_t|$ and $M=0$ in Algorithm \ref{alg:UCB}). In the forthcoming Section \ref{sec:unlabel}, we consider an extension where the DM starts with a non-empty set of feature data ${\cal S}_1 = \{(\theta_{h, -m})_{h=1}^H\}_{m\in [M]}$, where $(\theta_{h, -m})_{h=1}^H\sim \prod^H_{h=1}\Lambda_h$ for each $m\in [M]$. 
For estimating $\rho$, the DM collates the labeled data $(\theta_{h,s}, A_{h,s} ,Y_{h,s})$ for each $h\in [H], s\in [t-1]\setminus {\cal J}_t$ to form the dataset ${\cal D}_t$ (Line \ref{alg:collate_label}). In forming ${\cal D}_t$, we exclude time steps when $a_\nulll$ is taken, since $\rho(\theta, a_\nulll)=0$ is already known to the DM. 
At the end of each period $t$, we incorporate new data in either ${\cal D}_t$ (when $t\in [t]\setminus {\cal J}_{t+1}$) or ${\cal S}_t$ (when $t\in {\cal J}_{t+1}$), but not both. Only in the former case when $t\in [t]\setminus {\cal J}_{t+1}$, we update our estimates on $\rho$ (Line \ref{alg:Jt_static}) by applying the CB oracle ${\cal O}$ on ${\cal D}_{t+1}$, and update the estimates $\{\hat{U}_{h, t+1}\}_{h\in [H]}$ to $\{U_h\}_{h\in [H]}$ by invoking \textsf{Optimistic-DP} (Line \ref{alg:invoke}). Such a lazy update rule is for analysis purposes.


\begin{algorithm}[t]
	\caption{\textsf{Mimic-Opt-DP}}\label{alg:online}
	\begin{algorithmic}[1]
		\State \textbf{Input:} CB Oracle ${\cal O}$, $\delta > 0$, $\{{\cal J}_t\}^\infty_{t=1}$ ($=(\ref{eq:J})$ by default), ${\cal S}_1 $ ($=\emptyset$ by default).
		\State \textbf{Initialize: }${\cal D}_1 =\emptyset$,  $\text{UCB}_1(\theta, a;\delta) = 1, \text{LCB}_1(\theta, a;\theta) = 0$ $\forall\theta\in \Theta, a\in {\cal A}$, $\delta_t = \delta/t^2$.
        \State \textbf{Initialize: }$\hat{U}_{h, 1}(b)= 0$ for all $h\in [H], b\in \{0\}\cup [LH].$
		\For{$t=1,\ldots, T$}
			\For{$h = 1, \ldots, H$}
				\State Observe current inventory $b_{h, t}$ ($b_{1, t} = B_t$), and feature $\theta_{h, t}\sim \Lambda_h$. 
				\State \label{alg:online_Aht} Select an $A_{h, t}\in {\cal A}(b_{h,t},\theta_{h,t})$ that solves
				\begin{align}
				&\max_{a\in {\cal A}(b_{h,t},\theta_{h,t})} \left\{\text{UCB}_t(\theta_{h,t}, a;\delta_t) r(\theta_{h,t}, a) \right.\nonumber\\
				&\qquad \left. + \text{LCB}_t(\theta_{h,t}, a;\delta_t) \hat{U}_{h+1,t}(b- d(\theta_{h,t}, a)) + \left[1-\text{LCB}_t(\theta_{h,t}, a;\delta_t)\right] \hat{U}_{h+1,t}(b)\right\}.\label{eq:alg_online_Aht}
				\end{align}
				\State Observes $Y_{h,t}\sim \text{Bern}(\rho(\theta_{h, t}, a_{h,t}))$.
                \State Earn reward $R_{h, t} = r(\theta_{h, t}, A_{h, t})\cdot Y_{h,t}$, update $b_{h+1, t} = b_{h,t} - d(\theta_{h, t}, A_{h, t})\cdot Y_{h,t}$.
			\EndFor
			\If{$t\in {\cal J}_{t+1}$} \label{alg:begin_split}
				\State Update ${\cal S}_{t+1} = {\cal S}_t\cup \{(\theta_{h, t})^H_{h=1}\}$.   \label{alg:collate_unlabel}
                \State Set ${\cal D}_{t+1} = {\cal D}_t$, $\text{UCB}_{t+1} = \text{UCB}_t$, $\text{LCB}_{t+1} = \text{LCB}_t$, $\hat{U}_{h, t+1} = \hat{U}_{h, t}~\forall h\in [H]$. 
			\Else \Comment{$t\in [t]\setminus {\cal J}_{t+1}$}
				\State Update ${\cal D}_{t+1}= {\cal D}_t\cup \{(\theta_{h, t}, A_{h, t}, Y_{h, t}): h\in [H], A_{h, t}\neq a_{\nulll}\}$. \label{alg:collate_label}
                \State Update $\text{UCB}_{t+1}, \text{LCB}_{t+1} = {\cal O}({\cal D}_{t+1}, \delta_{t+1})$, set ${\cal S}_{t+1} = {\cal S}_t$.\label{alg:Jt_static}
                \State \label{alg:invoke} Invoke \textsf{Optimistic-DP}($\text{UCB}_{t+1}, \text{LCB}_{t+1}, \delta_{t+1}, {\cal S}_{t+1}$), which returns $\{\hat{U}_{h,t+1}\}^H_{h=1}$.
			\EndIf \label{alg:end_split}
		\EndFor
	\end{algorithmic}
\end{algorithm}

\begin{algorithm}[t]
	\caption{\textsf{Optimistic-DP}($\text{UCB}_t$, $\text{LCB}_t$, $\delta_t$, ${\cal S}_t$)}\label{alg:UCB}
	\begin{algorithmic}[1]
		\State \textbf{Input: }$\text{UCB}_t, \text{LCB}_t: \Theta\times{\cal A}\times(0,1)\rightarrow [0,1]$, $\delta_t > 0$, ${\cal S}_t = \{(\theta_{h, s})_{h\in [H]}\}_{s\in {\cal J}_t}\cup {\cal S}_1$, where ${\cal S}_1 = \{(\theta_{h, -m})_{h\in [H]}\}_{m\in [M]}$ represents a possibly empty (i.e. $M=0$) initial dataset.
		\State \textbf{Initialize: }$\hat{U}_{H+1,t}(b) = 0$ for all $b\in \{0\}\cup [LH].$
		\For{$h = H, \ldots 1$}\label{alg:UCB_for_start}
            \State\label{alg:UCB_Uhat} Compute $\hat{U}_{h,t}(b) = \frac{1}{|{\cal J}_t| + M} \sum_{s\in {\cal J}_t\cup \{-1, \ldots, -M\}} \hat{V}_{h,t}(b, \theta_{h,s})\wedge (b \wedge (H-h+1))r_{\text{max}}$ for each $b\in \{0\}\cup [LH]$. The function $\hat{V}_{h,t}: (\{0\}\cup [LH])\times \Theta\rightarrow \mathbb{R}$ is defined as
			\begin{align}
				&\hat{V}_{h,t}(b, \theta) = \max_{a\in {\cal A}(b, \theta)} \left\{\text{UCB}_t(\theta, a;\delta_t) r(\theta_t, a) \right.\nonumber\\
				&\qquad \left. + \text{LCB}_t(\theta, a;\delta_t) \hat{U}_{h+1,t}(b- d(\theta, a)) + \left[1-\text{LCB}_t(\theta, a;\delta_t)\right] \hat{U}_{h+1,t}(b)\right\}.\label{eq:hatVt} 		
			\end{align}
    	\EndFor\label{alg:UCB_for_end}
		\State Output $\{\hat{U}_{h, t}\}_{h\in [H+1]}$.
	\end{algorithmic}
\end{algorithm}

\subsection{Main Results}\label{sec:main}
Our main result is a high probability regret bound on \textsf{Mimic-Opt-DP} displayed in Algorithm \ref{alg:online}.
\begin{theorem}\label{thm:main}
Consider \textsf{Mimic-Opt-DP} with CB oracle ${\cal O}, \delta\in (0, 1)$, $\{{\cal J}_t\}^\infty_{t=1}$ specified in (\ref{eq:J}) and ${\cal S}_1 = \emptyset$. With probability at least $1-9\delta$, it holds that $\text{Reg}(T) \leq $
\begin{align}
    & \tilde{O}\left(r_\text{max}L \cdot H \sqrt{T}\right) + O\left( r_\text{max}L\sum_{t\in [T]\setminus{\cal J}_{T+1}} \sum^H_{h=1} [\text{UCB}_t(\theta_{h, t},A_{h,t}) - \text{LCB}_t(\theta_{h, t},A_{h,t})] \right), \nonumber
\end{align}   
where $\tilde{O}(\cdot)$ hides a multiplicative factor of absolute constant times $\sqrt{\log(LHT/\delta)}$, and $O(\cdot)$ hides a multiplicative factor of absolute constant. In particular, the regret bound is independent of $|\Theta|$.
\end{theorem}
Theorem \ref{thm:main} is proved in Section \ref{sec:highlevel}. In the regret bound, $r_\text{max} L$ is an upper bound on the opportunity cost of resource consumption (of at most $L$ units) in a time step, see Lemma \ref{lem:bound}. The first term is due to learning $\{\Lambda_h\}^H_{h=1}$. This term is \emph{independent of} $|\Theta|$. Instead of learning each $\Pr(\theta_h=\theta)$, we estimate its influence on the value-to-go functions in $\textsf{Optimistic-DP}$, leading to the saving. 

The second term is due to estimating $\rho$ via the CB oracle ${\cal O}$. Recall that $[T]\setminus {\cal J}_{T+1}$ consists of the episodes when we invoke ${\cal O}$. Thus, $(\text{Loss}) = \sum_{t\in [T]\setminus{\cal J}_{T+1}} \sum^H_{h=1} [\text{UCB}_t(\theta_{h, t},A_{h,t}) - \text{LCB}_t(\theta_{h, t},A_{h,t})] $ reflects how fast the confidence radius under ${\cal O}$ shrinks, when we update the UCBs, LCBs episode by episode, and each episode generates $\leq H$ samples (Line \ref{alg:collate_label} in Algorithm \ref{alg:online}). As surveyed in Appendix \ref{app:oracles}, existing research works on batched contextual bandits \cite{RenZK22} imply that $(\text{Loss}) = o(T)$ and independent of $|\Theta|$ for models such as contextual generalized linear bandit model \cite{FilippiCGS10}, contextual logistic bandit model \cite{FauryACF2020} and contextual Gaussian process bandit model \cite{KrauseO11}. 

To further illustrate, we specialize Theorem \ref{thm:main} on some of the applications in Table \ref{tab:application}. The specialized regret bounds are displayed in the ``$M = 0, {\cal S}_1 = \emptyset$'' column in Table \ref{tab:reg}, where we consider the applications with dynamic pricing (see Appendix \ref{app:pricing}) and first price auctions  (see Appendix \ref{app:auction}). More applications can be found and their regret bounds are in Appendix \ref{app:application}. In the first row on dynamic pricing with generalized linear bandit model, the parameters $\ell_f,\kappa_f>0$ are bounding parameters related to a demand-curve function $f$ that models $\rho$, see Appendix \ref{app:pricing}.

\begin{table}[h]
  \centering
  \resizebox{\columnwidth}{!}{%
  \begin{tabular}{|c||c|c|}
    \hline
     & $M=0$, ${\cal S}_1 = \emptyset$ & $M >0$, ${\cal S}_1 = \{(\theta_{h, -m}\}_{m\in [M]}$ \\
    \hline
    Dynamic pricing with &  $\tilde{O}(r_\text{max} H\sqrt{T}$
        & With $M\geq HT\kappa_f / (d \ell_f)$, \\
    gen. linear model & $+ r_\text{max} d \cdot \frac{\ell_f}{\kappa_f}\sqrt{HT} + r_\text{max} dH) $& $\tilde{O}(r_\text{max} d \cdot \frac{\ell_f}{\kappa_f}\sqrt{HT} + r_\text{max} dH) $\\
    \hline
    First price auction on  &    $\tilde{O}(r_\text{max}L H\sqrt{T}$
    & With  $M\geq HT/|{\cal A}|$, \\
    identical items & $ + r_\text{max} L\sqrt{|{\cal A}|HT} + r_\text{max}L |{\cal A}|H)$  & $\tilde{O}(r_\text{max}L \sqrt{|{\cal A}|HT} + r_\text{max}L |{\cal A}|H)$\\
    \hline
    First price auction on & & For any $M>0$, \\
    distinct items & $\tilde{O}(r_\text{max}L H\sqrt{|{\cal A}|T} + r_\text{max}L |{\cal A}|H^2)$&$\tilde{O}(r_\text{max}L H\sqrt{|{\cal A}|T} + r_\text{max} L |{\cal A}|H^2)$ \\
    \hline
  \end{tabular}%
}
  \caption{Regret bounds of $\text{Reg}(T)$ for some applications (details in Appendix \ref{app:application})}
  \label{tab:reg}
\end{table}

We compare Theorem \ref{thm:main} with existing works on contextual BwKs \cite{BadanidiyuruKS18,AgrawalDL16,AgrawalD16,LiS22,HanZWXZ23,SlivkinsZSF24,GuoL24,ChenAYPWD24,GuoL25,LiZ23}, which focus on the case of $T=1$ and achieve $\text{Reg}(1) = \textsf{opt}_1 - \sum^H_{h=1} R_{h, 1} = o(H)$. Our model and results have three advantages and one shortcoming. The three advantages are: (1) we allow $B_t$ to be any positive integer, while existing works require $B_1 = \Omega(H^c)$ for some $c\in [1/2,1]$, (2) we allow $\{\theta_{h, t}\}^H_{h=1}$ to be differently distributed across $1\leq h\leq H$, different from these existing works that require $\{\theta_{h, t}\}^H_{h=1}$ to be independently and identically distributed, (3) our results achieve $\text{Reg}(T) = o(T)$, which is not implied by existing works, as explained in the next paragraph. The shortcoming is we restrict to only one resource type, while existing works allow multiple resource types. While our framework allows the generalization of $\text{Reg}(T) = o(T)$ to $K>1$ resource settings, the resulting algorithm is computationally intractable. The generalization requires computing $\hat{U}_{h, t}(b)$ for $H(LH)^K$ many values of $b$, the remaining array of the $K$ resource types. Such intractability remains even if $\rho, \{\Lambda_h\}_{h\in [H]}$ are known.

We explain on (3): existing results on contextual BwK do not imply $\text{Reg}(T) = o(T)$. Firstly, with an unknown non-stationary $\{\Lambda_h\}^H_{h=1}$, it is impossible to achieve $\text{Reg}(1) = o(H)$ even if $\rho$ is known (Theorem 4 in \cite{JiangLZ25}). Thus, the existing results of $\text{Reg}(1) = o(H)$ on contextual BwKs crucially require $\Lambda_1 = \ldots = \Lambda_H$. Instead of considering $\textsf{opt}_1$ directly, which is intractable for general $K$, these existing results consider a fluid relaxation, leading to a linear program whose optimum $\textsf{UB}_1$ satisfies $\textsf{UB}_1 \geq \textsf{opt}_1$ (see Appendix \ref{app:supp_CBwK} for the precise definition of $\textsf{UB}_1$). Subsequently, these works focus on showing $\textsf{UB}_1 -  \sum^H_{h=1} R_{h, 1} = o(H)$ for a variety of contextual BwK settings. While it is tempting to generalize their results to $\sum^T_{t=1}[\textsf{UB}_t -  \sum^H_{h=1} R_{h, t}] = o(T)$, such an aim is impossible. Indeed, Proposition 4 in \cite{BumpensantiW20} demonstrates a one-episode instance with $B_1 = \Omega(H)$ such that $\textsf{UB}_1 - \textsf{opt}_1 = \Omega(\sqrt{H})$, meaning that any online algorithm must suffer $\Omega(\sqrt{H}T)$ regret when comparing to the fluid relaxation benchmark $\sum^T_{t=1}\textsf{UB}_t$.

\subsection{A High Level View on the Proof of Theorem \ref{thm:main}}\label{sec:highlevel}
We start with the following Lemma \ref{lem:bound} and Corollary \ref{cor:bound_V}, which are crucial for controlling how the estimation errors on the value-to-go function propagate in the backward \textbf{for} loop in \textsf{Optimistic-DP}.
\begin{lemma}\label{lem:bound}
For any given $t$, the inequality
\begin{equation}\label{eq:bound}
0\leq \hat{U}_{h, t}(b) - \hat{U}_{h, t}(b-d) \leq 2 r_\text{max} L
\end{equation}
holds for every $h\in [H], t\in [T], b\in [LH], d \in [\min\{b, L\}]$, and any realization of $\text{UCB}_t, \text{LCB}_t$ with certainty. In particular, for $U_h(b) = \mathbb{E}_{\theta_h\sim \Lambda_h} \left[V_h(b, \theta_h)\right]$, the inequality $0\leq U_h(b) - U_h(b-d) \leq 2 r_\text{max} L$ holds for every $h\in [H], t\in [T], b\in [LH], d \in [\min\{b, L\}]$.
\end{lemma}
The Lemma is proved in Appendix \ref{app:pf_lemma_bound}. Directly from Lemma \ref{lem:bound} we have:
\begin{cor}\label{cor:bound_V}
For any $h\in [H], b\in \{0\}\cup [LH], t\in [T]$, and any $\theta, \theta'\in \Theta$, the bound $|\hat{V}_{h,t}(b, \theta) - \hat{V}_{h,t}(b, \theta') | \leq (2L+1)r_\text{max}$ holds with certainty. In addition, for $h\in [H], b\in \{0\}\cup [LH]$, and any $\theta, \theta'\in \Theta$, the bound $|V_{h}(b, \theta) - V_{h}(b, \theta') | \leq (2L+1)r_\text{max}$ holds.
\end{cor}
Corollary \ref{cor:bound_V} is crucial for bounding the one-sided error of $\hat{U}_{1, t}(B_t)$ on estimating $\textsf{opt}_t$.
\begin{lemma}\label{lemma:optimism}
Consider the event ${\cal F}_{t, \delta'} = \{ \textsf{opt}_t \leq \hat{U}_{1, t}(B_t) + (2L+1) r_\text{max} H\sqrt{\frac{\log(LH^2/\delta')}{|{\cal J}_t|+M}} \}$ for a fixed $t\in [T]$ and $\delta'\in (0,1)$. It holds that $\Pr({\cal F}_{t, \delta'})\geq 1-\delta_t - \delta'$.
\end{lemma}
The Lemma is proved in Appendix \ref{app:pf_lemma_optimism}. While Lemma \ref{lemma:optimism} holds for general ${\cal S}_1, M$, Theorem \ref{thm:main} only requires the case ${\cal S}_1 = \emptyset, M =0$. The Lemma facilitates our regret analysis. Define the event
\begin{equation}\label{eq:event_cb}
    {\cal G}_{t, \delta_t} = \left\{\rho(\theta, a)\in [\text{LCB}_t(\theta,a;\delta_t), \text{UCB}_t(\theta, a;\delta_t)] \text{ for all $\theta\in \Theta, a\in {\cal A}$}\right\}
\end{equation}
when the CB oracle returns a set of consistent $\text{UCB}_t, \text{LCB}_t$. We have $\Pr({\cal G}_{t, \delta_t})\geq 1-\delta_t.$ Now, we are ready to provide our main result in the following.
\begin{theorem}\label{thm:regt}
In an episode $t\in [T]$ in \textsf{Mimic-Opt-DP}, the regret at episode $t$ satisfies the following upper bound almost sure: $\textsf{opt}_t - \sum^H_{h=1}R_{h, t}\leq$ 
\begin{align}
     &  6r_\text{max}L\cdot H \sqrt{\frac{\log(2LH^2/\delta')}{|{\cal J}_t|+M}} + 3r_\text{max}L\sum^H_{h=1} [\text{UCB}_t(\theta_{h, t},A_{h,t}) - \text{LCB}_t(\theta_{h, t},A_{h,t})] \label{eq:main}\\
    + & r_\text{max} B_t\cdot \mathbf{1}(\neg {\cal F}_{t, \delta'}) + 3r_\text{max}L\cdot H \cdot\mathbf{1}(\neg {\cal G}_{t, \delta_t})+ 3r_\text{max} L \sum^H_{h=1}\mathbf{1}(\neg \tilde{\cal E}_{h,t, \delta'})+ (\star_t). \label{eq:regt}
\end{align}
The event ${\cal F}_{t, \delta'}$ is defined in Lemma \ref{lemma:optimism}, the event ${\cal G}_{t, \delta_t}$ is defined in (\ref{eq:event_cb}). The event $\tilde{\cal E}_{h,t, \delta'}$, which satisfies $\Pr(\tilde{\cal E}_{h,t, \delta'})\geq 1-\delta'/H$, is defined in (\ref{eq:event_E_tilde}) in Appendix \ref{app:pf_thm_regt}. The term $(\star_t)$ is a random variable with the properties that $\mathbb{E}[(\star_t)] = 0$, and $\Pr(\sum^T_{t=1}(\star_t) \leq 2r_\text{max}(2L+1) \sqrt{HT\log(2/\delta'')}) \geq 1-3\delta''$ for any $\delta''\in (0,1)$.
\end{theorem}
Theorem \ref{thm:regt} is proved in Appendix \ref{app:pf_thm_regt}. Similar to Lemma \ref{lemma:optimism}, Theorem \ref{thm:regt} holds for general ${\cal S}_1, M.$ The two terms in (\ref{eq:main}) are interpreted similarly to those in Theorem \ref{thm:main}, while the quantities in (\ref{eq:regt}) are remnant terms that capture low-probability events, except the term $(\star_t)$ which accounts for stochastic variations and can be handled by the Azuma-Hoeffding inequality. 

Now, we apply $\delta' = \delta / T$ and $\delta'' = \delta$ in (\ref{eq:regt}) in Theorem \ref{thm:regt} and sum over the upper bound (\ref{eq:regt}) from $t=1$ to $T$. Under $\{{\cal J}\}^\infty_{t=1}$ specified in (\ref{eq:J}), the following regret bound holds with probability $1-8\delta$:
\begin{align}
    &\text{Reg}(T) \leq  r_\text{max}(8L+4)H \left(\frac{2}{\alpha}-1\right)\sqrt{\alpha T\log\frac{2LH^2T}{\delta}} \label{eq:regret_by_J}\\
    + & r_\text{max}(2L+1)\sum^T_{t=1} \sum^H_{h=1} [\text{UCB}_t(\theta_{h, t},A_{h,t}) - \text{LCB}_t(\theta_{h, t},A_{h,t})] + 6r_\text{max}L \sqrt{HT\log\frac{2}{\delta}}. \label{eq:regT}
\end{align}
The upper bound (\ref{eq:regret_by_J}) is because consecutive elements in (\ref{eq:J}) in ${\cal J}_{T+1}$ are at most $(2/\alpha)-1$ apart, and $|{\cal J}_{T+1}|\leq \alpha T$. To show $\text{Reg}(T) = o(T)$, it suffices to show $\sum^T_{t=1} \sum^H_{h=1} [\text{UCB}_t(\theta_{h, t},A_{h,t}) - \text{LCB}_t(\theta_{h, t},A_{h,t})] = o(T)$. While such a bound appears similar to existing results on UCB-based algorithms, we caution that in our algorithm we only use the data in episodes indexed by $[t-1]\setminus {\cal J}_t$ to construct $\text{UCB}_t, \text{LCB}_t$, different from existing results where all data in episodes/rounds indexed by $[t-1]$ are used to construct the upper or lower confidence bounds at $t$. We reconcile the connection to existing results on UCB by the following:
\begin{lemma}\label{lem:final_AH}
Consider \textsf{Mimic-Opt-DP} with input index set $\{{\cal J}_t\}^\infty_{t=1}$ specified in (\ref{eq:J}). 
It holds that $\Pr(\sum^T_{t=1} \sum^H_{h=1} [\text{UCB}_t(\theta_{h, t},A_{h,t}) - \text{LCB}_t(\theta_{h, t},A_{h,t})] \leq 2\sum_{t\in [T]\setminus {\cal J}_{T+1}} \sum^H_{h=1} [\text{UCB}_t(\theta_{h, t},A_{h,t}) - \text{LCB}_t(\theta_{h, t},A_{h,t})]  + H(1+\sqrt{\alpha T\log(2/\delta)}))\geq 1-\delta$.
\end{lemma}
The Lemma, which uses the design in (\ref{eq:J}) and our lazy update in Lines \ref{alg:Jt_static}, \ref{alg:invoke} in Algorithm \ref{alg:online}, is proved in Appendix \ref{app:pf_final_AH}. The upper bound in Lemma \ref{lem:final_AH} allows us to use existing regret bounds in batched contextual bandits (see Appendix \ref{app:oracles}) to provide an upper bound to $\sum^T_{t=1} \sum^H_{h=1} [\text{UCB}_t(\theta_{h, t},A_{h,t}) - \text{LCB}_t(\theta_{h, t},A_{h,t})]$. Combining Lemma \ref{lem:final_AH} with (\ref{eq:regT}), Theorem \ref{thm:main} is proved.
\subsection{Benefits of Unlabeled Feature Data}\label{sec:unlabel}
We finally consider the case when the DM is provided with a set of unlabeled features ${\cal S} = \{(\theta_{h, -m})_{h\in [H]}\}_{m\in [M]}$, where $(\theta_{h,-m})_{h\in [H]}\sim \prod^H_{h=1}\Lambda_h$ for each $m\in [M]$, before the online process starts. The dataset ${\cal S}$ is unlabeled, since it does not contain the allocation decision $A$ or conversion realization $Y$ associated with any feature in ${\cal S}$. In a dynamic pricing setting in an e-commerce platform, an array $(\theta_{h,-m})_{h\in [H]}$ corresponds to the features of customers who visited the platform during an episode (say a day). Such feature data can be available even before the introduction of a new product or new posted prices. In the first price auction settings for internet ad allocations, an array in ${\cal S}$ corresponds to the DM's valuations on the impressions arriving in an episode, which can be accessed with past internet data, without knowing other bidders' valuations or bids.

\textsf{Mimic-Opt-DP} benefits from the availability of ${\cal S} = \{(\theta_{h, -m})_{h\in [H]}\}_{m\in [M]}$, and enjoys improved regret bounds in some settings. Consider the following index sets $\{\tilde{\cal J}_t\}^\infty_{t=1}$ parameterized by an absolute constant $\alpha\in (0,1)$ and a positive integer $n_\alpha$ satisfying $1/\alpha\leq n_\alpha < (2/\alpha) -1$:
\begin{equation}\label{eq:J_tilde}
\tilde{\cal J}_\infty = \{1 + (M+i)\cdot n_\alpha: i\in \mathbb{Z}_{\geq 0}\}, \quad \tilde{\cal J}_t = \tilde{\cal J}_\infty \cap [t-1] \text{ for each $t\in \mathbb{Z}_{>0}$}.
\end{equation}
By inputting $\{\tilde{\cal J}_t\}^\infty_{t=1}$ in (\ref{eq:J_tilde}) to \textsf{Mimic-Opt-DP}, we ensure that at the start of any episode $t$, there are at least $(1-\alpha)t$ episodes reserved for learning $\rho$, and there are at least $\max\{M, \alpha t/2\}$ many independent random arrays $\sim \prod^H_{h=1}\Lambda_h$ for learning $\{\Lambda_h\}^H_{h=1}$. By specializing $M=0$, (\ref{eq:J_tilde}) reduces to (\ref{eq:J}). By applying Theorem \ref{thm:regt} and Lemma \ref{lem:final_AH}, we have the following performance guarantee:
\begin{cor}\label{cor:unlabeled}
    Consider \textsf{Mimic-Opt-DP} with input being CB oracle ${\cal O}$, $\delta\in (0,1)$, $\{\tilde{\cal J
    }_t\}^\infty_{t=1}$ according to (\ref{eq:J_tilde}), and ${\cal S}_1 = \{(\theta_{h, -m})_{h\in [H]}\}_{m\in [M]}$. With probability $\geq 1-9\delta$, \textsf{Mimic-Opt-DP} satisfies the following regret upper bound: $\text{Reg}(T) \leq$
    \begin{align}
    &  \tilde{O}\left(r_\text{max}L \cdot \frac{HT}{\sqrt{\max\{M, T\}}}\right) + O\left( r_\text{max}L\sum_{t\in [T] \setminus \tilde{\cal J}_{T+1}} \sum^H_{h=1} [\text{UCB}_t(\theta_{h, t},A_{h,t}) - \text{LCB}_t(\theta_{h, t},A_{h,t})] \right).\nonumber
\end{align}  
\end{cor}
The improvement of Corollary \ref{cor:unlabeled} compared to Theorem \ref{thm:main} is in the first term in the regret bound, which improves from $H\sqrt{T}$ to $HT/\sqrt{\max\{M, T\}}$. The second terms in the regret bounds in Corollary \ref{cor:unlabeled} and Theorem \ref{thm:main} have the same order bound. Comparing Corollary \ref{cor:unlabeled} with Theorem \ref{thm:main}, there is a strict improvement with the former when $H\sqrt{T}\neq O(\sum_{t\in [T] \setminus {\cal J}_{T+1}} \sum^H_{h=1} [\text{UCB}_t(\theta_{h, t},A_{h,t}) - \text{LCB}_t(\theta_{h, t},A_{h,t})] )$, but $M$ is large enough so that $HT/\sqrt{M}= O(\sum_{t\in [T] \setminus {\cal J}_{T+1}} \sum^H_{h=1} [\text{UCB}_t(\theta_{h, t},A_{h,t}) - \text{LCB}_t(\theta_{h, t},A_{h,t})])$. Table \ref{tab:reg} showcases the improved regret bounds under the column ``$M >0$, ${\cal S}_1 = \{(\theta_{h, -m}\}_{m\in [M]}$''. There is a strict improvement for dynamic pricing and first price auction on identical items, when $M$ is large enough. For first price auction on distinct items, there are $H (|{\cal A}| - 1)$ unknown parameters on learning $\rho$. The number of latent parameters is too large, and dominates the saving by ${\cal S}$, hence no improvement.  

Corollary \ref{cor:unlabeled} has an interesting implication on efficiently approximating the optimum $\textsf{opt}$ in an episode, which has a fixed but arbitrary starting inventory $\bar{B}\in [LH]$. Let's assume (only in this paragraph) that the DM knows $\rho$, so that he can set $\text{UCB}_t = \text{LCB}_t = \rho$, and the DM possesses the unlabeled feature data ${\cal S} = \{(\theta_{h, -m})_{h\in [H]}\}_{m\in [M]}$, but still does not know $\{\Lambda_h\}^H_{h=1}$. Approximating $\textsf{opt}$ is still challenging due to the intractability of the Bellman equations (\ref{eq:bellman}). Consider an instance ${\cal I}$ with $M$ episode, where $B_t = \bar{B}$ for all $t\in [M]$. Let's run $\textsf{Mimic-Opt-DP}$ on instance ${\cal I}$ with input ${\cal S}_1 = {\cal S}$ and CB oracle knowing $\rho$. Corollary \ref{cor:unlabeled} guarantees that $\textsf{opt} - \frac{1}{M}\sum^M_{t=1} \sum_{h=1}^H R_{h,t} \leq \tilde{O}(r_\text{max} L H / \sqrt{M})$ with probability $\geq 1-9\delta$. Next, the Azuma-Hoeffding inequality ensures the latter inequality in $\textsf{opt} - \frac{1}{M}\sum^M_{t=1} \sum_{h=1}^HR_{h,t} \geq \mathbb{E}[\frac{1}{M}\sum^M_{t=1}\sum_{h=1}^H R_{h,t}] - \frac{1}{M}\sum^M_{t=1}\sum_{h=1}^H R_{h,t} \geq - \tilde{O}(r_\text{max} H / \sqrt{M})$ holds with probability $1-\delta$. Altogether, we have the approximation guarantee $|\textsf{opt} - \frac{1}{M}\sum^M_{t=1}\sum_{h=1}^H R_{h,t}| \leq \tilde{O}(r_\text{max} L H / \sqrt{M})$ which holds with probability $\geq 1-10\delta$, and the $\tilde{O}(\cdot)$ hides a multiplicative factor of absolute constant times $\sqrt{\log(LHM/\delta)}$ \emph{independent} of $|\Theta|$.

\printbibliography

\newpage
\appendix

\section{Supplemental Discussions}

\subsection{Discussions on the Model Applications}\label{app:application}

This Section provides the details on three applications of the model:
\begin{enumerate}
    \item Dynamic Pricing in Section \ref{app:pricing},
    \item Dynamic Procurement in Section \ref{app:procurement},
    \item First Price Auctions in Section \ref{app:auction},
\end{enumerate}
which are also tabulated in Table \ref{tab:application}. For each application, we first provide the detailed set-up. Then, we state an appropriate CB oracle to use contingent on the model on $\rho$. Lastly, we state the corresponding regret bounds under the specific CB oracles.

\subsubsection{Dynamic Pricing}\label{app:pricing}
The DM sells a perishable resource over $T$ repeated episodes. A perishable resource can be fresh produce, which has a short shelf life. An episode can be a day, for example. In inventory control settings, it is often that the starting inventory $B_t$ in each episode is controlled or affected by a third party different from the seller (the DM). For example, $B_t$ could be influenced by the amount of fresh produce from a farm in the previous day, or $B_t$ could be constant across $t$ if the fresh produce's supplier and the DM have an agreement on ensuring a stream of stable supplies in multiple episodes.

At the start of episode $t$, the DM is supplied with $B_t$ units of the resource. The DM sells the resource units by posting dynamically adjusted prices during the episode. At each of the time steps $(1, t), \ldots, (H, t)$, at most one customer (corresponding to a request in Section \ref{sec:model})  arrives. At time step $(h, t)$, the DM first observes the context $\theta_{h, t}$ of the arriving customer, where $\theta_{h,t} = \theta_\nulll$ represents no arrival. The context $\theta_{h, t}$ encodes the purchase characteristic of the customer. Next, the DM reviews $b_{h, t}$, the amount of existing inventory. If $b_{h, t}\geq 1$, the DM offers to sell a unit of the product at the posted price $A_{h, t}\in {\cal A}(b_{h, t}, \theta_{h, t}) =  {\cal A}$, where $A_{h,t}$ generally depends on $\theta_{h, t}$ and the past observations. Then, the customer decides to purchase with probability $\rho(\theta_{h, t}, A_{h, t})$, and not to purchase with probability $1 - \rho(\theta_{h, t}, A_{h, t})$. In the former case, a revenue of $A_{h, t}$ is collected as the reward, while one unit of resource is consumed, i.e. $b_{h+1, t} = b_{h,t} - 1$. Else, in the case of $b_{h, t} = 0$, the DM has no unit to sell, and he must reject an incoming customer by offering the no-purchase price $a_\nulll$ (${\cal A}(b_{h, t}, \theta_{h,t}) = \{a_{\nulll}\}$). At the end of episode $t$, the remaining $b_{H+1, t}\geq 0$ units perish (hence scrapped).

We elaborate on the modeling of purchase probability $\rho(\theta, a)$ of a customer with feature $\theta$ under the offered price $a$. We assume the feature space $\Theta$ to be a subset of $\mathbb{R}^{d-1}$ for some $d\geq 2$. In receiving a unit of the resource, the customer gains an expected utility of $ \bar{\mu}^\top \theta$, where $\bar{\mu}\in \mathbb{R}^{d-1}$ encodes the characteristic of the resource. In paying the price $a$, the customer suffers a dis-utility of $u_0\cdot a$, where $u_0 > 0$. Overall, the customer's net utility is $ \bar{\mu}^\top \theta - u_0 a$. Following the Operations Research literature, we consider the following models on $\rho$ in terms of the net utility:  
\begin{itemize}
    \item \emph{Linear demand model:} $\rho(\theta, a) = \bar{\mu}^\top \theta - u_0 a$. \cite{BanK21,LiZ23}, 
    \item \emph{Generalized linear demand model:} $\rho(\theta, a) = f(\bar{\mu}^\top \theta - u_0 a)$ \cite{WangTL24}, where $f$ is an inverse link function (see Section \ref{sec:glb}), and an example is the logistic function $f_\text{log}$.
\end{itemize}
Both the linear demand model and generalized linear demand model can be modeled by the generalized linear bandit model in Section \ref{sec:glb}, where we set $\mu_* = (\bar{\mu}, -u_0)$, and we set the feature vector $\phi(\theta, a) = (\theta, a)\in \mathbb{R}^{d}$.
Correspondingly, we have $r(\theta, a) = a$, and $d(\theta, a) = 1$ for any $\theta\in \Theta\setminus \{\theta_\nulll\}, a\in {\cal A}\setminus \{a_\nulll\}$, and we have $r(\theta, a) = d(\theta, a) = 0$ if $\theta = \theta_\nulll$ or $a = a_\nulll$.

For the linear demand model, we apply the CB oracle for contextual linear bandits in (\ref{eq:cb_glm}) with $\ell_f = \kappa_f=  1$ in Section \ref{sec:glb}. In the setting of Theorem \ref{thm:main}, we attain the regret bound
$$
\text{Reg}(T) = \tilde{O}\left( H\sqrt{T} + d\sqrt{HT} +dH  \right).
$$
For the generalized linear demand model, we apply the CB oracle for contextual generalized linear bandits in (\ref{eq:cb_glm})  in Section \ref{sec:glb}. In the setting of Theorem \ref{thm:main}, we attain the regret bound
$$
\text{Reg}(T) = \tilde{O}\left( H\sqrt{T} + \frac{\ell_fd}{\kappa_f}\sqrt{HT} +dH  \right),
$$
where $\ell_f$ is the Lipschitz continuity constant of $f$, and $\kappa_f$ is the curvature parameter of $f$ defined in (\ref{eq:kappa}).

\subsubsection{Dynamic Procurement}\label{app:procurement}
Our model captures the dynamic procurement problem in a repeated episode setting. We follow \cite{BadanidiyuruKS12} on an online dynamic procurement model with pricing decisions in an episode, enriched by a contextual generalization. Dynamic procurement formulation in an episode captures the scenario when the DM has a finite compensation budget. The DM spends the budget by paying sequentially arriving workers (corresponding to requests in our general model in Section \ref{sec:model}) to do a certain job, such as labeling data such as images and pieces of texts for generating training data, or answering domain-specific questions for training Large-Language Models. The application of dynamic procurement is primarily related to the domain of crowd-sourcing, hosted by platforms (where the DM posts the jobs to be done) such as Amazon Mechanical Turks, Clickworker and Fiverr. 

In an episode $t$, the DM starts with a budget of $B_t\in [LH]$. At time step $(h, t)$ where $h\in [H]$, a worker arrives at the crowd-sourcing platform. The worker is associated with the context $\theta_{h,t}$ which encodes the worker's skillset and the worker's dis-utility in taking a job on the platform. Observing the context $\theta_{h,t}$, the DM offers to pay a price of $A_{h,t}\in [L]\cup \{0\}$ to the worker to perform his job. If the worker accepts the offer (i.e. $Y_{h,t}=1)$), the DM earns a constant utility of $r(\theta_{h,t}, A_{h,t}) = 1$, but consumes $d(\theta_{h,t}, A_{h,t}) =A_{h,t}$ units from his budget. Otherwise ($Y_{h,t} =0$), the DM earns no reward but consumes none of his resources. Altogether, the DM's realized reward and resource consumption are $r(\theta_{h,t}, A_{h,t})Y_{h,t}$ and $d(\theta_{h,t}, A_{h,t})Y_{h,t}$ respectively. 
As stated in the main model, we assume that $Y_{h,t}\sim \text{Bern}(\rho (\theta_{h,t}, A_{h,t}) )$, where $\rho$ can be modeled by various context bandit models, who have a non-trivial CB oracle. Lastly, the DM's remaining budget is updated as $b_{h+1, t} = b_{h,t} - d(\theta_{h,t}, A_{h,t})Y_{h,t}$. At the end of an episode, the DM claws back the remaining budget (if any), and the DM could start the next episode with an arbitrary budget according to his need.

We elaborate on the modeling of acceptance probability $\rho(\theta, a)$ of a worker with feature $\theta$ under the offered price $a$, in a similar way to the discussions in Section \ref{app:pricing} on dynamic pricing. We assume the feature space $\Theta$ to be a subset of $\mathbb{R}^{d-1}$ for some $d\geq 2$. In accepting the job request task, the worker gains an expected dis-utility of $ \bar{\mu}^\top \theta$, where $\bar{\mu}\in \mathbb{R}^{d-1}$ encodes the characteristic of the job. In receiving the price $a$, the worker earns a utility of $u_0\cdot a$, where $u_0 > 0$. Overall, the customer's net utility is $u_0 a- \bar{\mu}^\top \theta$. Similar to Section \ref{app:pricing}, we consider the following models on $\rho$ in terms of the net utility:  
\begin{itemize}
    \item \emph{Linear demand model:} $\rho(\theta, a) = u_0 a 
- \bar{\mu}^\top \theta $,
    \item \emph{Generalized linear demand model:} $\rho(\theta, a) = f(u_0 a - \bar{\mu}^\top \theta)$ \cite{WangTL24}, where $f$ is an inverse link function (see Section \ref{sec:glb}), and an example is the logistic function $f_\text{log}$.
\end{itemize}
Both the linear demand model and generalized linear demand model can be modeled by the generalized linear bandit model in Section \ref{sec:glb}, where we set $\mu_* = (\bar{\mu}, -u_0)$, and we set the feature vector $\phi(\theta, a) = (\theta, a)\in \mathbb{R}^{d}$.
Correspondingly, we have $r(\theta, a) = 1$, and $d(\theta, a) = a\in [L]$ for any $\theta\in \Theta\setminus \{\theta_\nulll\}, a\in {\cal A}\setminus \{a_\nulll\}$, and we have $r(\theta, a) = d(\theta, a) = 0$ if $\theta = \theta_\nulll$ or $a = a_\nulll$.

For the linear demand model, we apply the CB oracle for contextual linear bandits in (\ref{eq:cb_glm}) with $\ell_f = \kappa_f=  1$ in Section \ref{sec:glb}. In the setting of Theorem \ref{thm:main}, we attain the regret bound
$$
\text{Reg}(T) = \tilde{O}\left( LH\sqrt{T} + Ld\sqrt{HT} + LdH  \right).
$$
For the generalized linear demand model, we apply the CB oracle for contextual generalized linear bandits in (\ref{eq:cb_glm})  in Section \ref{sec:glb}. In the setting of Theorem \ref{thm:main}, we attain the regret bound
$$
\text{Reg}(T) = \tilde{O}\left( LH\sqrt{T} + \frac{L\ell_fd}{\kappa_f}\sqrt{HT} + LdH  \right),
$$
where $\ell_f$ is the Lipschitz continuity constant of $f$, and $\kappa_f$ is the curvature parameter of $f$ defined in (\ref{eq:kappa}).

\subsubsection{First Price Auction with Finite Budgets in Repeated Episodes}\label{app:auction}
Our model also captures first price auctions in online settings, when the DM is a bidder in an online platform, such as online ad auction platform. In each time step, an indivisible item (such as a digital ad impression in online ad allocation setting) arrives at the platoform. Then, the DM and other bidders submit their bids to the item. After that, the platform collect the bids, allocates the item to the highest bidder, and charges the winning bidder his/her bid. In recent years, there is a major shift from second-price auctions to first-price auctions in the online advertisement display markets, due to the latter's enhanced transparency and fairness \cite{WangYDK23,HanWz25}.

The DM is a bidder, who participates in $H$ rounds of first-price auctions in each episode, for episodes $1, \ldots, T$. In episode $t$, he starts with a budget of $b_{1, t} = B_t$. At time step $(h, t)$ for $h\in [H]$, the DM sees a particular item, which corresponds to a request in our general model in Section \ref{sec:model}. Then, the DM observes his private value $\theta_{h, t}\sim \Lambda_h$ on the item, where $\theta_{h,t}\in \Theta = \mathbb{R}_{\geq 0}$. Based on the historical observations, his private value $\theta_{h, t}$ and his remaining budget $b_{h, t}$, the DM submits a bid $A_{h, t}\in {\cal A}(b_{h, t}, \theta_{h, t}) = \{0, 1, \ldots, b_{h, t}\wedge L\}$. Here, we assume that the set of all possible bids is ${\cal A} = \{0, 1, \ldots, L\}$, with the 0 bid being the option of not bidding, under which the DM will not receive any reward but will not consume any of his budget. We denote $C_{h, t}$ as the highest other bid (HOB), meaning the maximum bid of all other bidders. When the DM chooses $A_{h, t}$, he does not observe $C_{h, t}$.

After $A_{h, t}$ is chosen and $C_{h, t}$ is realized, the DM
observes the conversion outcome $Y_{h, t} = \mathbf{1}(A_{h, t}\geq C_{h,t})$. The case $Y_{h, t}=1$ means the DM wins the auction, and we assume tie-breaking favors the DM without loss of generality (and also following the literature \cite{WangYDK23,HanWz25}). The case $Y_{h, t}=0$ means the DM loses the auction. We emphasize that only $\mathbf{1}(A_{h, t}\geq C_{h,t})$, but not the realization of $C_{h, t}$, is observed by the DM after $A_{h, t}$ is chosen. Existing works also study richer forms of observations, such as full feedback where the DM can observe $C_{h, t}$ \cite{WangYDK23}, and one-sided feedback where the DM only observes $C_{h, t}$ if $A_{h, t} < C_{h, t}$. Deriving improved regret bounds under these settings are interesting directions, but we focus on the case of only observing $Y_{h, t}$ for our discussions.

The DM receives the realized net utility $(\theta_{h,t} - A_{h,t})\mathbf{1}(A_{h, t} \geq C_{h, t})$, but consumes $A_{h, t}\mathbf{1}(A_{h, t} \geq C_{h, t})$ units of the budget. Here, we model that $r(\theta, a) = \theta - a$ and $d(\theta, a) = a$. After that, the DM proceeds to time step $(h+1, t)$ with the remaining budget $b_{h+1, t} = b_{h, t} - d(\theta_{h, t}, A_{h, t}) Y_{h, t}$.
At the end of round $(H, t)$, any remaining ($b_{H+1, t}$ units) budget is clawed back to the DM. In the next episode $t+1$, the DM starts with a potentially different budget $B_{t+1}$, which depends on a wide variety of factors such as the DM's interest in the amount of advertising in the future.

We complete the model by specifying the model on $C_{h,t }$, which we model as a random variable following \cite{BalseiroGMMS19,HanWz25,WangYDK23}. The model on $C_{h,t }$ leads to the model on $\rho(\theta, a)$, and we also comment on how $r(\theta, a), d(\theta, a)$ in our general model map to the current first price auction application. We consider two settings:
\begin{itemize}
    \item \emph{Identical items. }For each $t$, $C_{1, t}, \ldots, C_{H,t}$ are independently and identically distributed (iid) with the common distribution ${\cal G}$. In this case, the $H$ items are identical in nature, thus the HOB are iid. For each $\theta\in \Theta, a\in {\cal A}\setminus \{a_\nulll\}$, we have $\rho(\theta, a) = \Pr_{C_{h, t}\sim {\cal G}}(a \geq C_{h, t})$, and we denote $\Pr_{C_{h, t}\sim {\cal G}}(a \geq C_{h, t}) = \mu(a)$, a latent parameter to be learned for each $a\in {\cal A}\setminus \{a_\nulll\}$ (also recall that $\rho(\theta, a_\nulll) = 0$ for any $\theta\in \Theta$). In addition, we have $r(\theta, a) = \theta- a$ and $d(\theta, a) = a$. Altogether, we have $Y_{h, t}\sim \text{Bern}(\mu(A_{h,t}))$. Additionally,  $r(\theta_{h, t}, A_{h, t})Y_{h, t}$  and $d(\theta_{h, t}, A_{h, t})Y_{h, t}$ respectively model the realized reward and consumption at time step $(h, t)$.
    \item \emph{Distinct items. }For each $t$, $C_{1, t}, \ldots, C_{H,t}$ are independently but differently distributed. For each $h\in [H]$, we have $C_{h, t}\sim {\cal G}_h$, thus the probability distribution of $C_{h, t}$ varies in different time step $h$ in an episode, but does not vary across different episode. This models the case when there is a cyclical patterns in the items arrivals across different episodes (say days), which is often the case in online ad allocation. Indeed, ad allocation platforms could see similar traffic patterns across different days, but within a day the arrival traffic is in general non-stationary.
    
   We denote $\Pr_{C_{h, t}\sim {\cal G}_h}(a \geq C_{h, t}) = \mu_h(a)$, a latent parameter to be learned for each $a\in {\cal A}\setminus \{a_\nulll\}, h\in [H]$. To connect the time varying conversion probability $\mu_h(a)$ to $\rho$, we embed the feature $\theta$ with its time step $h$ within an episode, to form a time-step dependent feature $(\theta,h)$  For each $\theta\in \Theta, h\in [H], a\in {\cal A}\setminus \{a_\nulll\}$, we have $\rho((\theta, h), a) = \Pr_{C_{h, t}\sim {\cal G}_h}(a \geq C_{h, t}) = \mu_h(a)$. Altogether, we have $Y_{h, t}\sim \text{Bern}(\rho((\theta_{h,t}, h),A_{h,t}))$. Additionally,  $r((\theta_{h, t}, h), A_{h, t})Y_{h, t}$  and $d((\theta_{h, t}, h), A_{h, t})Y_{h, t}$ respectively model the realized reward and consumption at time step $(h, t)$, where $r((\theta, h), a) = \theta - a, d((\theta, h), a) = a$.
\end{itemize}
Finally, we comment on the regret bound of \textsf{Mimic-Opt-DP} in the identical item setting and distinct item setting. In the identical item setting, the DM should apply the CB oracle for stationary conversion model (Equation \ref{eq:KarmCB} in Section \ref{sec:K}), and incurs a regret of $$\tilde{O}\left(r_\text{max}L H\sqrt{T} + r_\text{max}L \sqrt{|{\cal A}HT|} +r_\text{max}L |{\cal A}|H\right). $$  In the distinct item setting, the DM should apply the CB oracle for non-stationary conversion model (Equation \ref{eq:KarmCB_non-stat} in Section \ref{sec:K}), and incurs a regret of $$\tilde{O}\left(r_\text{max}L H\sqrt{T} + r_\text{max}L H \sqrt{|{\cal A}T|} +r_\text{max}L |{\cal A}|H^2\right).$$ Both bounds are also tabulated in Table \ref{tab:reg}.

\subsection{Examples of Confidence Bound (CB) Oracles}\label{app:oracles}
In what follows, we provide examples of CB oracles based on existing research works on multi-armed bandits. In each example, we first state the model on $\rho$, the conversion probability. Then, we construct a corresponding CB oracle that satisfies Assumption \ref{ass:conf}. Lastly, for the constructed CB oracle, we provide a regret upper bound on $\sum_{t\in [T]\setminus {\cal J}_{T+1}} \sum^H_{h=1} [\text{UCB}_t(\theta_{h, t},A_{h,t}) - \text{LCB}_t(\theta_{h, t},A_{h,t})]$, which represents the regret incurred by using the CB oracle ${\cal O}$ to estimate $\rho$ in the $T$ episodes, see the discussions in Section \ref{sec:main}. 
Except Section \ref{sec:K}, our discussion will involve a feature vector $\phi(\theta, a)$ associated with each $\theta, a$. All vectors in the section are column vectors in $\mathbb{R}^d$, thus the inner product between vectors $\phi, \mu$ is denoted as $\phi^\top \mu$. The $d\times d$ identity matrix is denoted as $I_d$.

\subsubsection{(Non-contextual) $(|{\cal A}| - 1)$-armed Bandits}\label{sec:K}
While we primarily focus on contextual models on the conversion probability $\rho(\theta, a)$, in certain applications such as first price auction (see Section \ref{sec:auction}) only the deterministic outcomes $r(\theta, a)$ and/or $d(\theta, a)$ depends on the context $\theta$, but not $\rho(\theta, a)$. In this case, the conversion probability $\rho(\theta, a)$ is modeled by a non-contextual $(A-1)$-armed bandit model.

\textbf{Models on $\rho.$} We propose to consider two models:
\begin{itemize}
\item \emph{Stationary conversion model.} For each $\theta\in \Theta, a\in {\cal A}\setminus\{a_\nulll\}$, we have $$\rho(\theta, a) = \mu(a)\in [0, 1].$$
\item \emph{Non-stationary conversion models}. We append the time step index $h$ to the feature $\theta$, so now the feature is $(\theta, h)$.  For each $(\theta, h)\in \Theta\times [H], a\in {\cal A}\setminus\{a_\nulll\}$, we have $$\rho((\theta, h), a) = \mu_h(a)\in [0, 1],$$ For each $a\in {\cal A}\setminus \{a_\nulll\}$, the sequence $\mu_1(a), \ldots, \mu_H(a)\in [0,1]$ can be arbitrarily varying, but different episodes share the same set of $\{\mu_h(a)\}_{h\in [H], a\in {\cal A}\setminus \{a_\nulll\}}$.
\end{itemize}

\textbf{CB oracle for Stationary Conversion Model.} Given input data ${\cal D} = \{(\theta_n, A_n, Y_n)\}^N_{n=1}$ and confidence parameter $\delta$, the oracle returns 
\begin{equation}\label{eq:KarmCB}
\text{UCB}(\theta, a, \delta) = \hat{\mu}(a) + \sqrt{\frac{\log(2N|{\cal A}| / \delta)}{\max\{m(a), 1\}}}, \quad \text{LCB}(\theta, a, \delta) = \hat{\mu}(a) - \sqrt{\frac{\log(2N|{\cal A}| / \delta)}{\max\{m(a), 1\}}},
\end{equation}
where
$$
m(a) = \sum^N_{n=1}\mathbf{1}(A_n = a), \quad \hat{\mu}(a) = \frac{\sum^N_{n=1} Y_n \mathbf{1}(A_n = a)}{\max\{m(a), 1\}}. 
$$

\textbf{CB oracle for Non-stationary Conversion Model.} Recall that in a non-stationary conversion model, the combined feature is now $(\theta, h), $ which is appended by the time step index $h$. Thus, the input dataset ${\cal D}$ takes the form $\{((\theta_n, h_n), A_n, Y_n)\}^N_{n=1}$. Given ${\cal D}, \delta$ the CB oracle returns for each $\theta\in\Theta, h\in [H], a\in {\cal A}\setminus \{a_\nulll\}$
\begin{equation}\label{eq:KarmCB_non-stat}
\text{UCB}((\theta, h), a, \delta) = \hat{\mu}_h(a) + \sqrt{\frac{\log(2NH|{\cal A}| / \delta)}{\max\{m_h(a), 1\}}}, \quad \text{LCB}((\theta, h), a, \delta) = \hat{\mu}_h(a) - \sqrt{\frac{\log(2NH|{\cal A}| / \delta)}{\max\{m_h(a), 1\}}},
\end{equation}
where
$$
m_h(a) = \sum^N_{n=1}\mathbf{1}(A_n = a, h_n = h), \quad \hat{\mu}_h(a) = \frac{\sum^N_{n=1} Y_n \mathbf{1}(A_n = a, h_n = h)}{\max\{m_h(a), 1\}}. 
$$

\textbf{Regret upper bounds. }In the stationary conversion model, the bound
\begin{align*}
&\sum_{t\in [T]\setminus {\cal J}_{T+1}} \sum^H_{h=1} [\text{UCB}_t(\theta_{h, t},A_{h,t}) - \text{LCB}_t(\theta_{h, t},A_{h,t})] \nonumber\\
\leq & H(|{\cal A}|-1) + 4\sqrt{\left(1 - \frac{\alpha}{2}\right) (|{\cal A}|-1)HT\log\left(\frac{2TH|{\cal A}|}{\delta}\right)}
\end{align*}
holds with certainty. 
In the non-stationary conversion model, the bound
\begin{align*}
&\sum_{t\in [T]\setminus {\cal J}_{T+1}} \sum^H_{h=1} [\text{UCB}_t((\theta_{h, t}, h),A_{h,t}) - \text{LCB}_t((\theta_{h, t}, h),A_{h,t})] \nonumber\\
\leq &  H^2(|{\cal A}|-1) + 4H\sqrt{\left(1 - \frac{\alpha}{2}\right) (|{\cal A}|-1)T\log\left(\frac{2TH^2|{\cal A}|}{\delta}\right)}
\end{align*}
holds with certainty.

\subsubsection{Contextual Generalized Linear Bandits \cite{FilippiCGS10,RenZK22}}\label{sec:glb}
In the forthcoming discussions, we follow \cite{RenZK22}, who provide a model, a set of confidence bounds and regret bounds in a batched contextual generalized linear bandit setting. Each $\theta\in \Theta, a\in {\cal A}\setminus \{a_\nulll\}$ is associated with a known feature vector $\phi(\theta, a)$, lying in $B(q):= \{\phi \in \mathbb{R}^d: \|\phi\|_2\leq q\}$. 

\textbf{Model on $\rho.$} For each $\theta\in \Theta, a\in {\cal A}\setminus\{a_{\nulll}\}$, we have $\rho(\theta, a) = f(\phi(\theta,a)^\top \mu_*)\in [0, 1]$. The function $f$ is known, but the parameter vector $\mu_*\in \mathbb{R}^d$ is not known. The DM only knows that $\mu_*\in B(1)$. Denote the domain of $f$ as ${\cal D} = [-q, q].$ Following the assumptions in \cite{LiLZ17,RenZK22}, the function $f: {\cal D} \rightarrow [0,1]$ is an \emph{inverse link function} satisfying the following properties:
\begin{itemize}
    \item $f:{\cal D} \rightarrow [0,1]$ is monotonically increasing,
    \item $f$ is Lipchitz continuous. There exists $\ell_f >0$ such that $|f(u) - f(v)|\leq \ell_f |u-v|$ for all $u, v\in{\cal D} $.
    \item $f$ is twice differentiable.
    \item It holds that 
    \begin{equation}\label{eq:kappa}
    \inf_{\mu, \phi: \|\mu - \mu_*\|_2\leq 2, \|\phi\|_2\leq q} f'(\phi^\top\mu)\geq \kappa_f > 0.
    \end{equation}
    \item It holds that $f''(u) \geq 0$ for all $u\in {\cal D}$.
\end{itemize}

Two common examples of $f$ are:
\begin{itemize}
    \item $f_\text{lin}(w) = w$, which reduces to the case of linear bandits \cite{AbbasiPS11}, and we have $\ell_f = \kappa_f = 1$.
    \item $f_\text{log}(w) = \frac{1}{1 + e^{-w}}$, which is also known as the logistic bandits. We have $\ell_f \leq 1/4 , \kappa_f \geq e^{-2q}$. 
\end{itemize}

\textbf{CB Oracle. }The oracle requires the knowledge of $T$, though the requirement can be relaxed by the doubling trick. The oracle inputs the dataset ${\cal D} = \{(\theta_n, A_n, Y_n)\}^N_{n=1}$ and confidence parameter $\delta$. Then, the oracle sets up the following parameters:
\begin{align}
    \lambda & = \frac{4d\log(1+T/d^2)}{\kappa^2_f},\nonumber\\
    \gamma_N &= \sqrt{\lambda} + \frac{1}{\kappa_f} \sqrt{2\log\left( \frac{1}{\delta}\right) + d\log\left( 1 + \frac{N}{d\lambda} \right)}, \label{eq:gamma_N}\\
    V &= \lambda I_d + \sum^N_{n=1}\mathbbm{1}(A_n\neq a_{\nulll})\phi(\theta_n, A_n)\phi(\theta_n, A_n)^\top. \nonumber
\end{align}
After that, the oracles solves for $\hat{\mu}$, the solution of the following equation:
\begin{equation*}
    \sum^N_{n=1} \mathbbm{1}(A_n\neq a_{\nulll})\left[ Y_n - f(\phi(\theta_n, A_n))\top \mu \right]\phi(\theta_n, A_n) = \kappa_f \lambda\mu.
\end{equation*}

The solving of the above is equivalent to solving a penalized maximum likelihood estimate. For a further illustration, we remark that in the special case of $f = f_\text{lin}$ \cite{AbbasiPS11}, the solution $\hat{\mu}$ is an $\ell^2$-regularized least squares estimate of a linear regression problem, and $\hat{\mu}$ takes the following closed form expression:
$$
\hat{\mu} = V^{-1}\left(\sum^N_{n=1} Y_n \phi(\theta_n, A_n)\right).
$$
where we consider $\phi(\theta, a_{\nulll})=0_d$ for all $\theta$ and $Y_n=0$ if $A_n=a_{\nulll}$. Then, we don't need to add the factor $\mathbbm{1}(A_n\neq a_{\nulll})$ in the summation. Now, for each $\phi\in \mathbb{R}^d$, define $\| \phi \|_{V^{-1}} = \phi^\top V^{-1} \phi$. The oracle output the following UCB, LCB for each $\theta\in \Theta, a\in {\cal A}\setminus \{a_\nulll\}$:
\begin{align}
    \text{UCB}(\theta, a) &=\min\left\{f(\phi(\theta, a)^\top \hat{\mu}) + \gamma_N \| \phi(\theta, a) \|_{V^{-1}}, 1\right\}, \nonumber\\
    \text{LCB}(\theta, a) &=\max\left\{f(\phi(\theta, a)^\top \hat{\mu}) - \gamma_N \| \phi(\theta, a) \|_{V^{-1}}, 0\right\}. \label{eq:cb_glm}
\end{align}
Lastly, for $a = a_\nulll$, we can set $\phi(\theta, a_\nulll) = 0_d$, the $d$-dimensional zero vector, so that $\text{UCB}(\theta, a_\nulll) = \text{LCB}(\theta, a_\nulll) =0$ for all $\theta\in \Theta$.

\textbf{Regret Upper Bound.} Define $V_t = \lambda I_d + \sum_{s\in [t-1]\setminus {\cal J}_{t}}\sum^H_{h=1}\phi(\theta_{h, s}, A_{h,s})\phi(\theta_{h,s}, A_{h,s})^\top$, where we recall that $\phi(\theta, a_\nulll)= 0_d$. By equation (5.2) in \cite{RenZK22}, we know that
\begin{align}
    \sum_{t\in [T]\setminus {\cal J}_{T+1}} \sum^H_{h=1} [\text{UCB}_t(\theta_{h, t},A_{h,t}) - \text{LCB}_t(\theta_{h, t},A_{h,t})] \leq 2\ell_f \cdot \gamma_{HT} \sum_{t\in [T]\setminus {\cal J}_{T+1}} \sum^H_{h=1} \| \phi(\theta_{h, t}, A_{h, t}) \|_{V^{-1}_t}.
\end{align}
Applying Lemma 5.2 in \cite{RenZK22} gives us the following
$$
\sum_{t\in [T]\setminus {\cal J}_{T+1}} \sum^H_{h=1} \| \phi(\theta_{h, t}, A_{h, t}) \|_{V^{-1}_t}\leq \sqrt{10}\left[\sqrt{\left(1-\frac{\alpha}{2}\right)dHT\log\left(\frac{\lambda+HT}{\lambda}\right)} + \frac{dH}{\sqrt{\lambda}}\log\left(\frac{\lambda+HT}{\lambda}\right)\right].
$$
Altogether, we have
\begin{align*}
    \sum_{t\in [T]\setminus {\cal J}_{T+1}} \sum^H_{h=1} [\text{UCB}_t(\theta_{h, t},A_{h,t}) - \text{LCB}_t(\theta_{h, t},A_{h,t})] \leq \tilde{O}\left(\frac{\ell_f\cdot d}{\kappa_f}\sqrt{\left(1-\frac{\alpha}{2}\right) HT} + dH\right),
\end{align*}
where $\tilde{O}(\cdot)$ hides a multiplicative factor of $\log^{3/2}(dTH/\kappa_f\delta)$.

\subsubsection{Refined Regret Bounds for Contextual Logistic Bandits 
 \cite{FauryACF2020,LiS22} }\label{sec:logistic}
In this section, we turn to focus on a special case of section \ref{sec:glb}, setting $f(w)=\frac{1}{1+e^{-w}}$. 

\textbf{Model on $\rho.$} Similar to the section \ref{sec:glb}, we take $\rho(\theta, a)= \frac{1}{1+e^{-\phi(\theta,a)^\top\mu_*}}$ for all $\theta\in \Theta, a\in\mathcal{A}\setminus{a_{\text{null}}}$. We still define $\kappa_f$ as $\inf_{\mu, \phi: \|\mu - \mu_*\|_2\leq 2, \|\phi\|_2\leq q} f'(\phi^\top\mu)\geq \kappa_f > 0$. 

\textbf{CB Oracle. }Knowledge of $T$ is not required in this case, while the oracle still take dataset ${\cal D} = \{(\theta_n, A_n, Y_n)\}^N_{n=1}$ as input, and confidence parameter $\delta$. Assuming $A_n\neq a_{\text{null}}$, the oracle solves for $\tilde{\mu}$.
\begin{align*}
    & \tilde{\mu}\in \arg\max_{\mu\in \mathbb{R}^d}\\
    & \sum_{n=1}^N\mathbbm{1}(A_n\neq a_\nulll)\left(Y_n \log f(\phi(\theta_n, A_n )^\top\mu_*)+(1-Y_n)\log(1-f(\phi(\theta_n, A_n )^\top\mu_*))\right)-\frac{\lambda}{2}\|\mu_*\|^2
\end{align*}
If $\tilde{\mu}\notin B(1)$, we further adopt a projection 
\begin{align*}
    \hat{\mu}\in\arg\min_{\mu\in B(1)}\|\psi(\mu)-\psi(\tilde{\mu})\|_{W_t(\theta)^{-1}}
\end{align*}
where 
\begin{align*}
    \psi(\mu)= & \lambda\mu+\sum_{n=1}^N \mathbbm{1}(A_n\neq a_\nulll)f(\phi(\theta_n, A_n )^\top\mu_*)\phi(\theta_n, A_n )\\
    W(\mu)= & \lambda I_d + \sum_{n=1}^N \mathbbm{1}(A_n\neq a_\nulll)f'(\phi(\theta_n, A_n )^\top\mu_*)\phi(\theta_n, A_n ) \phi(\theta_n, A_n )^\top.
\end{align*}
Function $f'(w)$ denote $\frac{d f(w)}{dw}=\frac{d \frac{1}{1+e^{-w}}}{dw}=\frac{e^{-w}}{(1+e^{-w})^2}$.

Given the above $\hat{\mu}$, we plug in the value and derive $\hat{P}(\theta, a) = f(\phi(\theta_n, A_n )^\top\hat{\mu})$ as the estimated conversion probability. Based on the calculation in Lemma 1 of \cite{LiS22}, and the fact that $\|B(1)\|_2=1$, we define 
\begin{align*}
    \epsilon(\theta, a) = \gamma_{N,\lambda, \delta} \sqrt{\frac{3}{2\kappa_f}}\|\phi(\theta, a)\|_{V^{-1}}
\end{align*}
where
\begin{align*}
    \gamma_{N,\lambda, \delta} = & \frac{3\sqrt{\lambda}}{2} + \frac{2}{\sqrt{\lambda}}\log\left(\frac{2^d}{\delta}\left(1+\frac{N}{4d\lambda}\right)^{\frac{d}{2}}\right)\\
    V = & \frac{\lambda}{\kappa_f} I_d + \sum_{n=1}^N \mathbbm{1}(A_n\neq a_\nulll) \phi(\theta_n, A_n ) \phi(\theta_n, A_n )^\top.
\end{align*}
Recall $d$ is the dimension of $\mu_*$. Then, the oracle is ready to output the following UCB, LCB for each $\theta\in \Theta, a\in {\cal A}\setminus \{a_\nulll\}$:
\begin{align}
    \text{UCB}(\theta, a) &=\min\left\{\hat{P}(\theta, a) + \epsilon(\theta, a), 1\right\}, \nonumber\\
    \text{LCB}(\theta, a) &=\max\left\{ \hat{P}(\theta, a) - \epsilon(\theta, a), 0\right\}. 
\end{align}
Similar to the section \ref{sec:glb}, we set $\text{UCB}(\theta, a_\nulll)$, $\text{LCB}(\theta, a_\nulll)$ both 0 for all context $\theta\in\Theta$.

\textbf{Regret Upper Bound.} Similar to section \ref{sec:glb}, define 
\begin{align*}
    \tilde{V}_t = & \frac{\lambda}{\kappa_f}I_d + \sum_{s\in [t-1]\setminus {\cal J}_{t}}\sum^H_{h=1}\mathbbm{1}(A_{h,s}\neq a_\nulll)\phi(\theta_{h, s}, A_{h,s})\phi(\theta_{h,s}, A_{h,s})^\top.
\end{align*}
From the above calculation, we get
\begin{align*}
    & \sum_{t\in [T]\setminus {\cal J}_{T+1}} \sum^H_{h=1} [\text{UCB}_t(\theta_{h, t},A_{h,t}) - \text{LCB}_t(\theta_{h, t},A_{h,t})]\\
    \leq & 2\sum_{t\in [T]\setminus {\cal J}_{T+1}} \sum^H_{h=1} \epsilon(\theta_{h,t},A_{h,t})\mathbbm{1}(A_{h,t}\neq a_\nulll)\\
    = & \gamma_{HT,\lambda,\delta}\sqrt{\frac{6}{\kappa_f}}\sum_{t\in [T]\setminus {\cal J}_{T+1}} \sum^H_{h=1} \|\phi(\theta_{h,t},A_{h,t})\|_{V_{t-1}^{-1}}\mathbbm{1}(A_{h,t}\neq a_\nulll)\\
    \leq & \gamma_{HT,\lambda,\delta}\sqrt{\frac{6}{\kappa_f}}\sum_{t\in [T]\setminus {\cal J}_{T+1}}\sqrt{H}\sqrt{\sum_{h=1}^H \phi(\theta_{h,t},A_{h,t})^\top  V_{t-1}^{-1} \phi(\theta_{h,t},A_{h,t})}
\end{align*}
The last inequality is guaranteed by the Cauchy inequality. By the Lemma 5.2 in the \cite{RenZK22}, we can conclude
\begin{align*}
    & \sum_{t\in [T]\setminus {\cal J}_{T+1}}\sqrt{\sum_{h=1}^H \phi(\theta_{h,t},A_{h,t})^\top  V_{t-1}^{-1} \phi(\theta_{h,t},A_{h,t})}\\
    \leq & \sqrt{10}\left(\sqrt{d(1-\frac{\alpha}{2})T\log\left(\frac{\lambda/\kappa_f + (1-\alpha/2)TH}{\lambda /\kappa_f}\right)}+\frac{d\sqrt{H}}{\sqrt{\lambda /\kappa_f}}\log\left(\frac{(1-\alpha/2)TH+\lambda /\kappa_f}{\lambda/\kappa_f}\right)\right),
\end{align*}
suggesting that
\begin{align*}
    & \sum_{t\in [T]\setminus {\cal J}_{T+1}} \sum^H_{h=1} [\text{UCB}_t(\theta_{h, t},A_{h,t}) - \text{LCB}_t(\theta_{h, t},A_{h,t})]\\
    \leq & \gamma_{HT,\lambda,\delta}\sqrt{60}\left(\sqrt{\frac{d}{\kappa_f}(1-\frac{\alpha}{2})HT\log\left(\frac{\lambda/\kappa_f + (1-\alpha/2)TH}{\lambda /\kappa_f}\right)}+\frac{dH}{\sqrt{\lambda}}\log\left(\frac{(1-\alpha/2)TH+\lambda/ \kappa_f}{\lambda/\kappa_f}\right)\right).
\end{align*}
Recall
\begin{align*}
    \gamma_{HT,\lambda,\delta} = & \frac{3\sqrt{\lambda}}{2} + \frac{2}{\sqrt{\lambda}}\log\left(\frac{2^d}{\delta}\left(1+\frac{HT}{4d\lambda}\right)^{\frac{d}{2}}\right)\\
    = & O\left(\sqrt{\lambda} + \frac{d+\log\frac{1}{\delta}}{\sqrt{\lambda}}+d\log\left(1+\frac{HT}{d\lambda}\right)\right)
\end{align*}
In summary, we can conclude
\begin{align*}
    \sum_{t\in [T]\setminus {\cal J}_{T+1}} \sum^H_{h=1} [\text{UCB}_t(\theta_{h, t},A_{h,t}) - \text{LCB}_t(\theta_{h, t},A_{h,t})]\leq \tilde{O}\left(d^{\frac{3}{2}}\sqrt{\frac{1}{\kappa_f}(1-\alpha/2)HT}+d^2H\right)
\end{align*}

where $\tilde{O}(\cdot)$ hides a multiplicative factor of $\log^{3/2}(dTH/\kappa_f \delta)$.

\subsubsection{Contextual Kernelized Bandits \cite{ValkoKMFC13,ChowdhuryG17}}
\label{sec:Kernel-Bandit}

\textbf{Model on $\rho$.} Generalizing \cite{ValkoKMFC13,ChowdhuryG17}, we consider the modeling of $\{\rho(\theta, a)\}_{\theta\in \Theta, a\in {\cal A}\setminus \{a_\nulll\}}$ in a non-Bayesian Reproducing Kernel Hilbert Space (RKHS) setting, which is a frequentist analogue to Gaussian processes, and provides a non-parameteric model on $\rho$ different from Sections \ref{sec:glb}. \ref{sec:logistic}. 
Firstly, we associate each context-action pair $(\theta, a)\in \Theta\times {\cal A} \setminus\{a_\nulll\}$ with a known feature vector $\phi(\theta, a)$ lying in a compact set $\Phi\subset \mathbb{R}^m$, and postulate that there is a function $p: \Phi \rightarrow [0,1]$ such that $p(\phi(\theta, a)) = \rho(\theta, a)$ for each $(\theta, a)\in \Theta\times {\cal A} \setminus\{a_\nulll\}$. \cite{KrauseO11} provides methodology on constructing such feature maps in contextual Gaussian process bandit settings, which can be adopted in the current kernelized bandit setting. Estimating $p(\phi(\theta, a))$ for all $(\theta, a)\in \Theta\times {\cal A} \setminus\{a_\nulll\}$ is intractable due to $\Theta$, which can be infinite in size. 

Following \cite{ValkoKMFC13,ChowdhuryG17} and being motivated by Gaussian processes, we assume that $p \in \Xi$, where $\Xi$ consists of functions that have a small RKHS norm. We provide a more precise definition of $\Xi$ in the following discussion. We let $\Xi$ be an RKHS of functions that map $\Phi \rightarrow [0,1]$. The RKHS $\Xi$ is associated with a positive semi-definite kernel function $k:\Phi\times\Phi\rightarrow \mathbb{R}$. The kernel $k$ induces an inner product operator $< \cdot, \cdot >_k$ that obeys the reproducing property: $\bar{p}(\phi) = <\bar{p}, k(\phi, \cdot) >_k$ for all $\bar{p}\in \Xi$ and all $\phi\in \Phi$, and that $k(\phi, \phi')\leq 1$ for all $\phi, \phi'\in \Phi$. Common examples of such kernels are the Squared Exponential kernel and the Matern kernel, respectively defined as
\begin{align*}
    k_\text{SE} (\phi,\phi') &= \exp\left( - \frac{\|\phi - \phi'\|^2_2}{2\ell^2}\right),\\
    k_\text{Matern} (\phi,\phi') & =\frac{2^{1-\nu}}{\Gamma(\nu)} \left(\frac{\|\phi - \phi'\|_2\sqrt{2\nu}}{\ell}\right)^\nu B_\nu\left( \frac{\|\phi - \phi'\|_2\sqrt{2\nu}}{\ell} \right),
\end{align*}
where $\ell>0, \nu>0$ are hyper-parameters, $\|\cdot\|_2$ is the Euclidean norm on $\mathbb{R}^m$, and $B_\nu(\cdot)$ is the modified Bessel function. 

We precisely define $\Xi= \{\bar{p}: \Phi \rightarrow [0,1] | \|\bar{p}\|_k \leq C \}$, where $C>0$ is a bounding constant and $\|\bar{p}\|_k = \sqrt{< \bar{p}, \bar{p} >_k}$. The condition $\|\bar{p}\|_k \leq C $ quantifies the smoothness of $\bar{p}$, in the sense that $|\bar{p} (\phi) -\bar{p}(\phi')| \leq \|\bar{p}\|_k  \| k(\phi,\cdot) - k(\phi', \cdot) \|_k$.  In our online setting, the DM does not know the exact identity of $\rho$ or $p$. However, the DM knows the feature map $\phi$, and the DM knows the bounding parameter $C$ and the kernel $k$ associated with $\Xi$. 

\textbf{CB oracle. }We derive a CB oracle based on \cite{ChowdhuryG17}. While \cite{ChowdhuryG17} considers a non-contextual kernelized bandit setting, they provide a powerful analytical tool of self-normalizing concentration inequality in the kernelized bandit setting, where the underlying RKHS can be infinite dimensional which makes the application of the self-normalized concentration inequality for linear bandits \cite{AbbasiPS11} invalid. 

The CB oracle inputs the dataset ${\cal D} = \{(\theta_n, A_n, Y_n)\}^N_{n=1}$ and the confidence parameter $\delta$. The oracle also has additional inputs being the regularizer $\lambda > 1$, the kernel $k$ and the bounding parameter $C$  for $\Xi$. Firstly, we compute the feature vector $\phi_n = \phi(\theta_n, A_n)$ for each $n\in [N]$. Secondly, we compute the kernel matrix on the observed data: $\hat{K} = [k(\phi, \phi')]_{\phi, \phi'\in \{\phi_1, \ldots, \phi_N\}}\in \mathbb{R}^{N\times N}$. Thirdly, we define the functions $\hat{k}:\Phi\rightarrow \mathbb{R}^N$, $\hat{\mu}:\Phi\rightarrow \mathbb{R}$, $\tilde{k}: \Phi\times\Phi\rightarrow \mathbb{R}$, $\hat{\sigma}:\Phi\rightarrow \mathbb{R}_{\geq 0}$:
\begin{align}
    \hat{k}(\phi) &= [k(\phi_1, \phi), \ldots, k(\phi_N, \phi)]^\top, \nonumber\\
    \hat{\mu}(\phi) & = \hat{k}(\phi)^\top (\hat{K} + \lambda I)^{-1} [Y_1, \ldots Y_N]^\top,\nonumber\\
    \tilde{k}(\phi, \phi')&=k(\phi,\phi') - \hat{k}(\phi)^\top(\hat{K} + \lambda I)^{-1}\hat{k}(\phi'),\nonumber\\
    \hat{\sigma}(\phi) &= \sqrt{\tilde{k}(\phi, \phi)},\label{eq:kernel_sigma}
\end{align}
where $I$ denotes the $N\times N$ identity matrix. 
Lastly, given any $(\theta, a) \in \Theta \times {\cal A} \setminus\{a_\nulll\}$, the oracle returns
\begin{align}
    \text{UCB}(\theta, a) & = \min \{ \hat{\mu}(\phi(\theta, a)) + \beta \hat{\sigma}(\phi(\theta, a)), 1\}\nonumber\\
     \text{LCB}(\theta, a) & = \max\{ \hat{\mu}(\phi(\theta, a)) - \beta \hat{\sigma}(\phi(\theta, a)) ,  0\},\nonumber
\end{align}
where 
\begin{equation}\label{eq:kernel_beta}
    \beta = C + \frac{1}{2}\cdot \sqrt{2 (\gamma_N + 1 + \log(1/\delta))},     
\end{equation}
and the quantity
\begin{equation*}
    \gamma_N = \max_{F\subset \Phi : |F|= N} I(Y_F; P_F).
\end{equation*}
Now, denote ${\cal N}(\mu, \Sigma)$ as the multi-variate Gaussian distribution with mean $\mu$ and covariance matrix $\Sigma$.
The quantity $I(Y_F; P_F)$ denotes the mutual information between $P_F = (P(\phi))_{\phi\in F}\sim {\cal N}(0_N, [k(\phi, \phi')]_{\phi, \phi'\in F})$ and $Y_F = F_F + \epsilon_F$, where $\epsilon_F\sim {\cal N}(0, \lambda I)$.  The mutual information $I(Y_F; P_F)$ quantifies the reduction in the uncertainty on $P_F$ after observing the noisy observations $Y_F$. The bounding parameter $\gamma_N$ is problem dependent, and in particular depends on $\Phi\subset \mathbb{R}^m$ and $k$. It is well-known that $\gamma_N = O((\log N)^{m+1})$ for the Squared Exponential kernel $k_\text{SE}$, and that $\gamma_N = O(N^{\frac{m(m+1)}{2\nu + m(m+1)}}\log N)$ for the Matern kernel. Lastly, if $a = a_\nulll$, we set $\text{UCB}(\theta, a) = \text{LCB}(\theta, a) = 0$.

By Theorem 2 in \cite{ChowdhuryG17} (our online sequence of feature vectors $(\phi(\theta_n, A_n))_{n\in [N] }$ corresponds to the online sequence of $x_1, \ldots, x_{t-1}$ in their Theorem 2), it holds that
$$
\Pr\left(\rho(\theta, a)= p(\phi(\theta, a)) \in [\text{LCB}(\theta, a), \text{UCB}(\theta, a)] \text{ for all $(\theta, a)\in \Theta \times {\cal A}\setminus \{a_\nulll\} $}\right) \geq 1-\delta. 
$$
Thus the above CB oracle satisfies Assumption \ref{ass:conf}.

\textbf{Regret Upper Bound. }We assert that a regret upper bound can be readily obtained by the analysis in \cite{ChowdhuryG17}, and the submodularity of log determinant on positive definite matrices (a well-known result, see \cite{kleinberg2019submodular} for example) for handling the regret bound in a batched setting. To start, we let $\beta_t, \sigma_t$ denote the quantities $\beta, \hat{\sigma}$ (see (\ref{eq:kernel_beta}, \ref{eq:kernel_sigma})) computed at the start of episode $t$, using the data in episodes $\{s: s\in [T]\setminus {\cal J}_{T+1}, s<t\}$. We claim that
\begin{align}
    &\sum_{t\in [T]\setminus {\cal J}_{T+1}} \sum^H_{h=1} [\text{UCB}_t(\theta_{h, t},A_{h,t}) - \text{LCB}_t(\theta_{h, t},A_{h,t})]\nonumber\\
    \leq & 2 \sum_{t\in [T]\setminus {\cal J}_{T+1}} \sum^H_{h=1} \beta_t \sigma_t(\phi(\theta_{h, t}, A_{h,t}))\cdot \mathbf{1}(A_{h,t}\neq a_\nulll)\nonumber\\
    \leq &2\left(C+\frac{1}{2}\sqrt{2\gamma_{HT} + 1 + \log(1/\delta)}\right)\cdot H\cdot \sqrt{4\lambda \gamma_T T}.\label{eq:kernel_main}
\end{align}
Apart from the factor $H$, the bound (\ref{eq:kernel_main}) is similar (in terms of order bound) to \cite{ChowdhuryG17}. 
We prove (\ref{eq:kernel_main}) in what follows. Firstly, since $\gamma_N$ is non-decreasing in $N$, we know that $\beta_t\leq \beta_T$ for all $t\in [T]$, thus
$$
\sum_{t\in [T]\setminus {\cal J}_{T+1}} \sum^H_{h=1} \beta_t \sigma_t(\phi(\theta_{h, t}, A_{h,t}))\cdot \mathbf{1}(A_{h,t}\neq a_\nulll) \leq \beta_T \sum_{t\in [T]\setminus {\cal J}_{T+1}} \sum^H_{h=1}\sigma_t(\phi(\theta_{h, t}, A_{h,t}))\cdot \mathbf{1}(A_{h,t}\neq a_\nulll).
$$
Next, we claim that, for any $h\in [H]$, it holds that
\begin{equation}\label{eq:kernel_claim}
    \sum_{t\in [T]\setminus {\cal J}_{T+1}} \sigma_t(\phi(\theta_{h, t}, A_{h,t}))\cdot \mathbf{1}(A_{h,t}\neq a_\nulll) \leq \sqrt{4\lambda \gamma_T T}.
\end{equation}
Proving (\ref{eq:kernel_claim}) leads to the proof of the regret upper bound (\ref{eq:kernel_main}), thus we focus on proving (\ref{eq:kernel_claim}) in the remaining. Now, the proof of Lemma 3 in \cite{ChowdhuryG17} implies that
\begin{align*}
& \sum_{t\in [T]\setminus {\cal J}_{T+1}} \sigma_t(\phi(\theta_{h, t}, A_{h,t}))\cdot \mathbf{1}(A_{h,t}\neq a_\nulll)\nonumber\\
 \leq &\sqrt{4\lambda T \cdot \frac{1}{2} \sum_{t\in [T]\setminus {\cal J}_{T+1}} \log(1+ \lambda^{-1} \sigma^2_t(\phi(\theta_{h, t}, A_{h,t}) ))\cdot \mathbf{1}(A_{h,t}\neq a_\nulll) },
\end{align*}
which crucially requires $\lambda > 1$. Next we analyze the term $\frac{1}{2}\cdot \log(1+ \lambda^{-1} \sigma^2_t(\phi(\theta_{h, t}, A_{h,t}) ))$, where $t\in [T]\setminus {\cal J}_{T+1}$ and $A_{h, t}\neq a_\nulll$. To this end, it is useful to recall that to construct $\sigma_t$ we use feature vector data $(\phi(\theta_{i, s}, A_{i, s}))_{(i, s)\in {\cal I}_{[H], <t}  }$, where the index set  ${\cal I}_{[H], <t}  = \{(i,s): s\in [t-1]\setminus {\cal J}_t, i\in [H], A_{i,s}\neq a_\nulll\}$ records the time steps when we take non-null action and the corresponding outcome is used for learning $\rho$. We further denote ${\cal I}_{[H], <t}^+ = {\cal I}_{[H], <t} \cup \{(h,t)\}$. To ease the remaining notation, we denote $\phi_{h, t} = \phi(\theta_{h, t}, A_{h,t})$ for each $h, t$. Consider the multi-variate Gaussian random variable $$(P_{i, s})_{(i, s)\in {\cal I}_{[H], <t}^+}  \sim {\cal N}(0, [k(\phi_{i, s}, \phi_{i', s'})]_{(i, s), (i', s')  \in {\cal I}_{[H], <t}^+  }) ,$$ and for each $(i, s)\in  {\cal I}_{[H], <t}^+$ we denote $\tilde{Y}_{i,s} = P_{i,s} + \epsilon_{i, s}$, where $\epsilon_{i,s}\sim {\cal N}(0, \lambda)$. The key observation is that
\begin{align}
    &\frac{1}{2}\cdot \log(1+ \lambda^{-1} \sigma^2_t(\phi(\theta_{h, t}, A_{h,t}) )) + \frac{1}{2}\cdot \log(2\pi e \lambda)\nonumber\\
     = &H(\tilde{Y}_{h, t} | (\tilde{Y}_{i, s})_{(i,s)\in {\cal I}_{[H], <t}}  ). \label{eq:cond_diff_entro}
\end{align}
In (\ref{eq:cond_diff_entro}), we follow the information theory literature on the notation of conditional differential entropy\footnote{At the risk of notational clash with $H$ being the length of an episode} $H(X_1 | X_2, \ldots, X_N) = H(X_1, X_2, \ldots X_N) - H(X_2, \ldots X_N)$, and $H(X_1, X_2, \ldots X_N)$ denotes the joint differential entropy of the continuous random variables $X_1, \ldots, X_N$. Consider ${\cal I}_{h, <t} =  \{(h,s): s\in [t-1]\setminus {\cal J}_t, A_{h,s}\neq a_\nulll\}$. Note that ${\cal I}_{h, <t}\subseteq {\cal I}_{[H], <t}$. The submodularity result in example 10 in \cite{kleinberg2019submodular} implies that the following inequality holds
$$
H(\tilde{Y}_{h, t} | (\tilde{Y}_{i, s})_{(i,s)\in {\cal I}_{[H], <t}}  )\leq H(\tilde{Y}_{h, t} | (\tilde{Y}_{i, s})_{(i,s)\in {\cal I}_{h, <t}}  ). 
$$
Finally, summing (\ref{eq:cond_diff_entro}) over $t\in [T]\setminus{\cal J}_{T+1}$ where $A_{h, t}\neq a_\nulll$, applying the above inequality and the chain rule, we get
\begin{align}
&\sum_{t\in [T]\setminus {\cal J}_{T+1}} \frac{1}{2}\cdot \log(1+ \lambda^{-1} \sigma^2_t(\phi(\theta_{h, t}, A_{h,t}) )) \cdot \mathbf{1}(A_{h, t}\neq a_\nulll) \nonumber\\
\leq & I( (\tilde{Y}_{h, t})_{(h,t)\in {\cal I}_{h, <T+1}}; (P_{h, t})_{(h,t)\in {\cal I}_{h, <T+1}}) \leq \gamma_T. \nonumber
\end{align}
Altogether, the desired regret upper bound is shown.

\subsubsection{Contextual Gaussian Process Bandits \cite{KrauseO11}}\label{sec:gauss}
\cite{KrauseO11} proposes a Gaussian Process Bandits problem, assuming the unknown reward function $f:\Theta \times \mathcal{A}\rightarrow \mathbb{R}$ is a sample from Gaussian Process $GP(0, k)$. For any $(\theta, a), (\theta', a')\in \Theta\times \mathcal{A}, a,a'\neq a_\nulll$, they assume $\mathbb{E}[f(\theta, a)] = 0$ and $k((\theta, a), (\theta', a')) = \mathbb{E}\left[f(\theta, a)f(\theta', a')\right]$. Here $k: \Theta\times \mathcal{A} \times \Theta\times \mathcal{A}\rightarrow \mathbb{R}$ is called Kernel function for a Gaussian Process. If we further assume the noise $\epsilon_n: = Y_n- f(\theta_n,A_n)$ follows Gaussian distribution $N(0, \sigma^2)$ for all $A_n\neq a_\nulll$, we can explicitly derive the posterior distribution for $f$ and concentration event. As we required $\rho(\theta, a)\in[0,1]$ and $Y_n$ follows Bernoulli distribution, we cannot directly assume $\rho$ is sampled from a Gaussian Process. But here we discuss the regret bound in the case of episode round setting, assuming the existence of an Oracle similar to the \cite{KrauseO11}.

\textbf{Model on $\rho$ and CB Oracle} Suppose $\Theta, \mathcal{A}$ are both finite. We do not impose any assumption on $\rho(\theta, a)$. Instead, we imitate the Theorem 4 in \cite{KrauseO11} to assume the existence of an Oracle, to approximate the $\rho(\theta, a)$.  We assume there exists an Oracle $\mathcal{O}$, which is able to output $\mu_{N}(\theta, a)$ and $\sigma_{N}(\theta, a)$ for all $\theta\in \Theta$ and $a\in \mathcal{A}, a\neq a_\nulll$, given dataset $\{Y_n, \theta_n, A_n\}_{n=1}^{N-1}$. And the output $\mu_{N}(\theta, a)$ and $\sigma_{N}(\theta, a)$ satisfy that
\begin{align*}
    & \Pr\left(\forall t\in\mathbb{N}, \forall \theta\in \Theta, a\in \mathcal{A}, |\mu_{t-1}(\theta, a)-\rho(\theta, a)| \leq \sqrt{\beta_t}\sigma_{t-1}(\theta, a)\right)\geq 1-\delta,\\
    & \sum_{t=1}^T\sigma^2_{t}(\theta_t,a_t) \leq \frac{2\gamma_T}{\log(1+1/\sigma^2)},\forall T\in \mathbb{N}.
\end{align*}
where $\beta_t = 2\log\left(|\mathcal{A}|\cdot|\Theta|t^2\pi^2/6\delta\right)$ and $\gamma_T$ is the maximum information gain within $T$ actions $\gamma_T:=\max_{A\subset \Theta\times\mathcal{A}: |A|=T} I(Y_A;\rho)$, the same as the definition in section \ref{sec:Kernel-Bandit}. $\sigma^2$ can be considered a proxy of noise level.

These assumptions hold if $\rho$ is a sample from Gaussian Process, with Gaussian noise $N(0, \sigma^2)$. Given the kernel function $k: \Theta\times \mathcal{A} \times \Theta\times \mathcal{A}\rightarrow \mathbb{R}$, $k(\theta, a; \theta', a')\leq 1$ holds for all $\theta, a; \theta', a'$. Given $\{(\theta_n, A_n, Y_n)\}^N_{n=1}$, we can define following notations as section \ref{sec:Kernel-Bandit}.
\begin{align*}
    \hat{k}(\theta, a) &= \Big[k\Big((\theta, a), (\theta_1,A_1)\Big), \ldots, k\Big((\theta, a), (\theta_N,A_N)\Big)\Big]^\top, \\
    \hat{K}& = \Big[k\Big((\theta_i, A_i), (\theta_j, A_j)\Big)\Big]_{i,j\in [T-1]}\\
    \tilde{k}\Big((\theta, a), (\theta', a')\Big)&=k\Big((\theta, a), (\theta', a')\Big) - \hat{k}\Big(\theta, a\Big)^\top(\hat{K} + \lambda I)^{-1}\hat{k}\Big(\theta', a'\Big),\\
    \hat{\sigma}(\phi) &= \sqrt{\tilde{k}(\phi, \phi)}\\
    \hat{\mu}(\phi) & = \hat{k}(\phi)^\top (\hat{K} + \lambda I)^{-1} [Y_1, \ldots ,Y_N]^\top,
\end{align*}
Theorem 4, 5 in \cite{KrauseO11} proves the above $\hat{\sigma}, \hat{\mu}$ are the targeted Oracle.

Given the guarantee of the Oracle, it is natural to define
\begin{align*}
    \text{UCB}_t(\theta, a) &=\min\left\{\mu_{t-1}(\theta, a) + \sqrt{\beta_t}\sigma_{t-1}(\theta, a), 1\right\}, \nonumber\\
    \text{LCB}_t(\theta, a) &=\max\left\{\mu_{t-1}(\theta, a) - \sqrt{\beta_t}\sigma_{t-1}(\theta, a), 0\right\}.
\end{align*}
at episode $t$, using all the data collected before $t$.

\textbf{Regret Upper Bound.} The proof for the upper bound of $\sum_{t\in [T]\setminus {\cal J}_{T+1}} \sum^H_{h=1} [\text{UCB}_t(\theta_{h,t}, a_{h,t}) - \text{LCB}_t(\theta_{h,t}, a_{h,t})]$ is very similar to the section \ref{sec:Kernel-Bandit}. By the calculation, we know $\gamma_t$ is not decreasing in $t\in\mathbb{N}$. 
\begin{align*}
    & \sum_{t\in [T]\setminus {\cal J}_{T+1}} \sum^H_{h=1} [\text{UCB}_t(\theta_{h,t}, a_{h,t}) - \text{LCB}_t(\theta_{h,t}, a_{h,t})]\\
    \leq & \sum^H_{h=1}\sum_{t\in [T]\setminus {\cal J}_{T+1}}   2\sqrt{\beta_{(t-1)H}}\sigma_{t-1}(\theta_{h,t}, a_{h,t})\\
    \leq & 2\sqrt{\beta_{TH}}\sum^H_{h=1}\sqrt{| [T]\setminus {\cal J}_{T+1}|}\sqrt{\sum_{t\in [T]\setminus {\cal J}_{T+1}}\sigma_{t-1}^2(\theta_{h,t}, a_{h,t})}\\
    \leq & 2\sqrt{\beta_{TH}} \cdot H \sqrt{T}\cdot\sqrt{\frac{\gamma_{HT}}{\log(1+1/\sigma^2)}}\\
    \leq & O\left(H\sqrt{\frac{T \gamma_T}{\log(1+1/\sigma^2)} \log\left(\frac{|\mathcal{A}|\cdot|\Theta|(TH)^2\pi^2}{6\delta}\right)}\right).
\end{align*}

\subsubsection{Contextual Neural Bandits}
In this section, we introduce \cite{pmlr-v119-zhou20a}, which address a $K$-armed contextual bandit problem by a neural network. \cite{pmlr-v119-zhou20a} assumes the context $\{x_{t,a}\}_{a=1}^K \in \mathbb{R}^{d\times K}$ sequentially arrives at each round $t\in [T] $, assuming $\|x_{t,a}\|_2=1$(Assumption 4.2 in \cite{pmlr-v119-zhou20a}). 

\textbf{Model on $\rho.$} For each $\theta\in \Theta, a\in {\cal A}\setminus\{a_{\nulll}\}$, we have $\rho(\theta, a) \in [0, 1]$. For simplicity and being consistent with the notation in \cite{pmlr-v119-zhou20a}, we concatenate the observed context and each action $a\in\mathcal{A}$, define $x_{t,h,a}:= [\theta_{h,t}, a]$ for each $h\in[H], t\in [T], a\in\mathcal{A}$. Or we can use one-hot encoder to encode $a$. Without loss of generality, we assume $x_{h,t,a}\in \mathbb{R}^d, \|x_{h,t,a}\|_2=1$ and $(x_{h,t,a})_j =(x_{h,t,a})_{j+d/2} $. As the section 4 in \cite{pmlr-v119-zhou20a} discusses, this assumption is mild. Denote $\{x^i\}_i^{TH|\mathcal{A}|}$ as a collection of all the contexts. In this paper, $H$ denotes the length of each episode. To avoid notation confusion, we turn to denote $\mathcal{H}$ as the neural tangent kernel matrix, while still define vector $\vec{h}=\{\rho(x^i)\}_i^{TH|\mathcal{A}|}$, following the \cite{pmlr-v119-zhou20a}. There is no restriction on the structure $\rho(x)$, but we assume a constant $S,\lambda_0$ satisfying $S\geq \sqrt{2h^\top\mathcal{H} h}$, $\mathcal{H}\succeq \lambda_0 I$ is known.

\textbf{CB oracle. } Given depth parameter $L\geq 2$, we take a large enough integer $m:=\text{poly}(T, H, |\mathcal{A}|, \frac{1}{\lambda}, \frac{1}{S}, \frac{1}{\lambda_0}, \log\frac{1}{\delta})$ as the width and appropriate $J,\eta$ as gradient descent step number and step size. The exact value of $m,J,\eta$ refer to Lemma 4.5 and Remark 4.7 in \cite{pmlr-v119-zhou20a}.  The overall idea is to use a neural network $f(x;\theta) = \sqrt{m} W_L\sigma(W_{L-1}\sigma(\cdots \sigma(W_1 x)))$ to approximate $\rho$. Here $x$ is the observed context and denote $\theta=[\text{vec}(W_1)^T, \cdots, \text{vec}(W_l)^T]^T$ is the weight to be updated. In the following section, $\theta$ is no longer the observed context but the weights in the neural network. Not hard to see the dimension of $\theta$ is $m+md + m^2(L-1)=:p$. We take $g(x;\theta)=\frac{\partial f}{\partial \theta}\in\mathbb{R}^p$ as the gradient of the weight.

We adopt the same hypermeter setting just as \cite{pmlr-v119-zhou20a}, such regularization parameter $\lambda$, step size $\eta$, coefficient $\{\gamma_{s}\}_{s=1}^{TH}$ gradient descent step number $J$ and call the algorithm 2 in \cite{pmlr-v119-zhou20a} to train the neural network. At the end of each episode $t\in [t]\setminus J_{t+1}$, we define the 
\begin{align*}
    Z_{t}=\lambda I_p + \sum_{t'\in [t]\setminus J_{t+1} }\sum_{h'=1}^{H}g(x_{h',t',a_{h',t'}})g(x_{h',t',a_{h',t'}})^\top
\end{align*}
and call the algorithm 2 in \cite{pmlr-v119-zhou20a} to update the neural network. Denote the updated parameter as $\theta_t$. We ignore the factor $\mathbbm{1}(a_{h',t'}\neq a_{\nulll})$, as we can take vector $g$ as a zero vector and skip the summation in this round, if we take action $a_{\nulll}$.

Since the algorithm 2 in \cite{pmlr-v119-zhou20a} always starts the training with the same initialization $\theta_0$, the theoretical analysis still holds when we feed multiple but at most $H$ new samples to the neural network. As the Lemma 5.1, 5.2 in \cite{pmlr-v119-zhou20a} suggest, an appropriate distribution of $\theta_0$ can guarantee with probability at least $1-\delta$, there exists $\theta^*\in\mathbb{R}^p$ such that
\begin{align}
    f(x^j; \theta_0)= & 0 \label{eq:neural-init-0}\\
    \rho(x^i)= & g(x^i;\theta_0)^\top(\theta^*-\theta_0)\label{eq:neural-true-rho-formulation}\\
    \|\theta^*-\theta_0\|_2\leq & \frac{2\sqrt{h^\top \mathcal{H}^{-1} h}}{\sqrt{m}}\label{eq:neural-theta*-theta0}\\
    \|\theta_t-\theta_0\|_2 \leq & \frac{2\sqrt{tH}}{\sqrt{m\lambda}}\label{eq:neural-theta_t-theta0}.
\end{align}
The last line is by the fact that there are at most $tH$ pieces of $(\theta_i, a_i, Y_i)$ being used to estimate $\theta_t$. Conditioned on the above inequality and the Lemma B.4, B.5, B.6, there exists 
$\bar{C}_3$, 
for any $\delta\in (0,1)$, following inequalities holds for all $j\in [TH|\mathcal{A}|]$ with probability $1-\delta$ 
\begin{align}
    |f(x^j; \theta_{t})-f(x^j; \theta_0)-g(x^j; \theta_0)^T(\theta_{t}-\theta_0)|\leq & \bar{C}_3\left(2\sqrt{\frac{tH}{m\lambda}}\right)^{\frac{4}{3}}L^3\sqrt{m\log m}\label{eq:neural-x-theta-gap}\\
    \|g(x^j; \theta_0)-g(x^j; \theta_{t})|_2\leq & \bar{C}_3\sqrt{\log m} (2\sqrt{\frac{tH}{m\lambda}})^{\frac{1}{3}} L^3 \|g(x^j; \theta_0)\|_2\label{eq:neural-gradient-gap}\\ 
    \|g(x^j; \theta_0)\|_2 \leq & \bar{C}_3\sqrt{mL}.\label{eq:neural-gradient-norm}
\end{align}
What's more, \cite{AbbasiPS11} concludes $\|\theta_t-\theta^*\|_{Z_{t-1}}\leq \frac{\gamma_{tH}}{\sqrt{m}}$ holds with probability $1-\delta$. 

Given the above inequalities, we are ready to bound the gap $|f(x^j; \theta_{t})-h(x^j)|$ for all $j\in [TH|\mathcal{A}|]$. The first step is to use triangle inequality to split the value into several parts.
\begin{align*}
    & |f(x^j; \theta_{t})-h(x^j)|\\
    \leq & | f(x^j; \theta_{t})-f(x^j; \theta_0)-g(x^j; \theta_0)^\top(\theta_{t}-\theta_0)|+\\
    & |(g(x^j; \theta_0)-g(x^j; \theta_{t}))^\top(\theta^*-\theta_{0})|+\\
    & |(g(x^j; \theta_0)-g(x^j; \theta_{t}))^\top(\theta_0-\theta_{t})|+|g(x^j; \theta_{t})^\top(\theta^*-\theta_{t})|.
\end{align*}
holds for all $j\in [TH|\mathcal{A}|]$. Notice that (\ref{eq:neural-init-0}) implies we can add $f(x^j; \theta_0)$ without deducting it. By (\ref{eq:neural-x-theta-gap}), we have
\begin{align*}
    | f(x^j; \theta_{t})-f(x^j; \theta_0)-g(x^j; \theta_0)^T(\theta_{t}-\theta_0)| \leq \bar{C}_3\left(2\sqrt{\frac{tH}{m\lambda}}\right)^{\frac{4}{3}}L^3\sqrt{m\log m}.
\end{align*}
By (\ref{eq:neural-theta*-theta0}), (\ref{eq:neural-gradient-gap}), (\ref{eq:neural-gradient-norm}), we have
\begin{align*}
    & |(g(x^j; \theta_0)-g(x^j; \theta_{t}))^\top(\theta^*-\theta_{0})|\\
    \leq & \|g(x^j; \theta_0)-g(x^j; \theta_{t})\|_2\|\theta^*-\theta_{0}\|_2\\
    \leq & \bar{C}_3\sqrt{\log m} (2\sqrt{\frac{tH}{m\lambda}})^{\frac{1}{3}} L^3 \bar{C}_3\sqrt{mL} \frac{\sqrt{h^T\mathcal{H}^{-1} h}}{\sqrt{m}}.
\end{align*}
By (\ref{eq:neural-theta_t-theta0}), (\ref{eq:neural-gradient-gap}), (\ref{eq:neural-gradient-norm}), we have
\begin{align*}
    & (g(x^j; \theta_0)-g(x^j; \theta_{t}))^\top(\theta_0-\theta_{t})\\
    \leq & \|g(x^j; \theta_0)-g(x^j; \theta_{t})\|_2 \|\theta_0-\theta_{t}\|_2\\
    \leq & \bar{C}_3\sqrt{\log m} (2\sqrt{\frac{tH}{m\lambda}})^{\frac{1}{3}} L^3 \bar{C}_3\sqrt{mL}\cdot 2\sqrt{\frac{tH}{m\lambda}}.
\end{align*}
By (\ref{eq:neural-gradient-norm}) and the conclusion $\|\theta_t-\theta^*\|_{Z_{t}}\leq \frac{\gamma_{tH}}{\sqrt{m}}$ from \cite{AbbasiPS11}
\begin{align*}
    & |g(x^j; \theta_{t})^\top(\theta^*-\theta_{t})|\\
    \leq & \|g(x^j; \theta_{t})\|_{Z_{t}^{-1}}\|\theta^*-\theta_{t}\|_{Z_{t}}\\
    \leq & \frac{\gamma_{tH}}{\sqrt{m}}\sqrt{g(x^j; \theta_{t})^{T}Z_{t}^{-1}g(x^j; \theta_{t}) }.
\end{align*}

In summary, we have probability at least $1-\delta$, such that
\begin{align*}
& |f(x_{h, t+1, a}; \theta_{t})-\rho(x_{h, t+1, a})|\\
\leq & \bar{C}_3\left(2\sqrt{\frac{tH}{m\lambda}}\right)^{\frac{4}{3}}L^3\sqrt{m\log m} + \frac{\gamma_{tH}}{\sqrt{m}}\sqrt{g(x_{h, t+1, a}; \theta_{t})^{T}Z_{t}^{-1}g(x_{h, t+1, a}; \theta_{t}) }+\\
& \bar{C}_3\sqrt{\log m} (2\sqrt{\frac{tH}{m\lambda}})^{\frac{1}{3}} L^3 \bar{C}_3\sqrt{mL}\left(\frac{\sqrt{h^T\mathcal{H}^{-1} h}}{\sqrt{m}} + 2\sqrt{\frac{tH}{m\lambda}}\right)\\
= & \bar{C}_3 \frac{(tH)^{\frac{2}{3}}}{m^{\frac{1}{6}}\lambda^{\frac{1}{6}}} (2)^{\frac{4}{3}}L^3\sqrt{\log m} + \frac{\gamma_{tH}}{\sqrt{m}}\sqrt{g(x_{h, t+1, a}; \theta_{t})^{T}Z_{t-1}^{-1}g(x_{h, t+1, a}; \theta_{t}) }+\\
& \bar{C}_3\sqrt{\log m}2^{\frac{1}{3}} \frac{(tH)^{\frac{1}{6}}}{m^{\frac{1}{6}}\lambda^{\frac{1}{6}}} L^3 \bar{C}_3\sqrt{L}\left(\sqrt{h^T\mathcal{H}^{-1} h} + 2\sqrt{\frac{tH}{\lambda}}\right)\\
\leq & \bar{C}_3 \frac{(tH)^{\frac{2}{3}}}{m^{\frac{1}{6}}\lambda^{\frac{1}{6}}} (2)^{\frac{4}{3}}L^3\sqrt{\log m} + \frac{\gamma_{tH}}{\sqrt{m}}\sqrt{g(x_{h, t+1, a}; \theta_{t})^{T}Z_{t}^{-1}g(x_{h, t+1, a}; \theta_{t}) }+\\
& \bar{C}_3\sqrt{\log m}2^{\frac{1}{3}} \frac{(tH)^{\frac{1}{6}}}{m^{\frac{1}{6}}\lambda^{\frac{1}{6}}} L^3 \bar{C}_3\sqrt{L}\left(S + 2\sqrt{\frac{tH}{\lambda}}\right)
\end{align*}
holds for all $h\in[H]$.

Take $\epsilon_t(x_{h, t+1, a})$ as the right handside of the last inequality, we can define 
\begin{align}
    \text{UCB}(x_{h, t+1, a}) & = \min \{ f(x_{h, t+1, a};\theta_t) + \epsilon_t(x_{h, t+1, a}), 1\}\nonumber\\
     \text{LCB}(x_{h, t+1, a}) & = \max\{ f(x_{h, t+1, a};\theta_t) - \epsilon_t(x_{h, t+1, a}) ,  0\},\nonumber
\end{align}
Though the theoretical analysis requires sufficiently large width $m$ and step number $J$, further a large radius function $\epsilon_t(x_{h, t+1, a})$, \cite{pmlr-v119-zhou20a} presents a much simple way to identify relatively good $m,J,\eta,\gamma_t$ value in practice. As the section 7 in \cite{pmlr-v119-zhou20a} suggests, conducting grid search on $\gamma_t=\gamma\in \{0.01,0.1,1,10\}$, $\lambda\in\{0.1, 1, 10\}$, $S\in\{0.01,0.1,1,10\}$ is sufficient to guarantee good numeric performance.

\textbf{Regret Upper Bound. } From the definition, easy to see
\begin{align*}
    & \sum_{t\in [T]\setminus {\cal J}_{T+1}} \sum^H_{h=1} [\text{UCB}_t(x_{h,t,a_t}) - \text{LCB}_t(x_{h,t,a_t})]\\
    \leq & H\underbrace{\sum_{t\in [T]\setminus {\cal J}_{T+1}} \bar{C}_3 \frac{(tH)^{\frac{2}{3}}}{m^{\frac{1}{6}}\lambda^{\frac{1}{6}}} (2)^{\frac{4}{3}}L^3\sqrt{\log m} + \bar{C}_3\sqrt{\log m}2^{\frac{1}{3}} \frac{(tH)^{\frac{1}{6}}}{m^{\frac{1}{6}}\lambda^{\frac{1}{6}}} L^3 \bar{C}_3\sqrt{L}\left(S + 2\sqrt{\frac{tH}{\lambda}}\right)}_{\dagger}+\\
    & \gamma_{TH}\underbrace{\sum_{t\in [T]\setminus {\cal J}_{T+1}} \sum^H_{h=1}\min\left\{\frac{1}{\sqrt{m}}\sqrt{g(x_{h, t, a}; \theta_{t-1})^{T}Z_{t-1}^{-1}g(x_{h, t, a}; \theta_{t-1}) }, 1\right\}}_{\ddagger}.
\end{align*}
We can upper bound $\dagger$ and $\gamma_{TH}$ by taking an appropriate width $m$, just as \cite{pmlr-v119-zhou20a}. Unlike section \ref{sec:logistic}, \ref{sec:glb}, we do not apply the Lemma 5.2 in the \cite{RenZK22}, as the dimension $d$ of $g(x;\theta)$ can be very large. Instead, we 
only conduct a loose analysis, finding an upper bound for each $h\in[H]$.

Define $Z_{h,t}=\lambda I_p + \sum_{t'=1}^{t}g(x_{h,t',a_{h,t'}}; \theta_{t-1})g(x_{h,t',a_{h,t'}}; \theta_{t-1})^\top$. Easy to see $Z_{t}\succeq Z_{t,h}$. Also, we imitate \cite{pmlr-v119-zhou20a} to define "effective dimension" for each $h$, which is $\tilde{d}_h = \frac{\log \text{det}(I+\mathcal{H}_h/\lambda)}{\log (1+T|\mathcal{A}|/\lambda)}$ and $\tilde{d}=\frac{\log \text{det}(I+\mathcal{H}/\lambda)}{\log (1+T|\mathcal{A}|/\lambda)}$, where $\mathcal{H}_h$ is the neural tangent kernel matrix defined by only using the $\{x_{h,t,a}\}_{t,a}$. Following the proof of lemma 5.4 in \cite{pmlr-v119-zhou20a}, we have
\begin{align*}
    & \sum_{t\in [T]\setminus {\cal J}_{T+1}} \min\left\{g(x_{h, t, a}; \theta_{t-1})^{T}Z_{t-1}^{-1}g(x_{h, t, a}; \theta_{t-1}) / m, 1\right\}\\
    \leq & \sum_{t\in [T]\setminus {\cal J}_{T+1}} \min\left\{g(x_{h, t, a}; \theta_{t-1})^{T}Z_{h,t-1}^{-1}g(x_{h, t, a}; \theta_{t-1}) / m, 1\right\}\\
    \leq & 2\log\frac{\text{det}( Z_{h, T})}{\text{det} (\lambda I)}\\
    \leq & C_1m^{-\frac{1}{6}}\sqrt{\log m}L^4T^{\frac{5}{3}}\lambda^{-\frac{1}{6}}+\left(\tilde{d}_h\log(1+\frac{T|\mathcal{A}|}{\lambda})+1\right).
\end{align*}
By the Cauchy inequality, we have
\begin{align*}
    & \sum_{t\in [T]\setminus {\cal J}_{T+1}} \sum^H_{h=1}\min\left\{\frac{1}{\sqrt{m}}\sqrt{g(x_{h, t, a}; \theta_{t-1})^{T}Z_{t-1}^{-1}g(x_{h, t, a}; \theta_{t-1}) }, 1\right\}\\
    \leq & \sum^H_{h=1} \sqrt{|[T]\setminus {\cal J}_{T+1}|}\sqrt{C_1m^{-\frac{1}{6}}\sqrt{\log m}L^4T^{\frac{5}{3}}\lambda^{-\frac{1}{6}}+ \left(\tilde{d}_h\log(1+\frac{|[T]\setminus {\cal J}_{T+1}|\cdot|\mathcal{A}|}{\lambda})+1\right)}
\end{align*}

In summary, by taking appropriate value of $m,J,\eta$, we have
\begin{align*}
    & \sum_{t=1}^T \frac{t^{\frac{2}{3}+\frac{1}{2}}H^{\frac{5}{3}+\frac{1}{2}} L^\frac{7}{2} \sqrt{\log m}}{ m^{\frac{1}{6}}\lambda^{\frac{5}{3}+\frac{1}{2}}} = O(1),\\
    & \gamma_{TH} = O(1)+O\left(\sqrt{1+\log\frac{1}{\delta} + \log\frac{\text{det}(Z_T)}{\text{det}(\lambda I)}}+\sqrt{\lambda}S\right).
\end{align*}
Plugin the above results, we can conclude with probability at least $1-\delta$, we have
\begin{align*}
    & \sum_{t\in [T]\setminus {\cal J}_{T+1}} \sum^H_{h=1} [\text{UCB}_t(x_{h,t,a_t}) - \text{LCB}_t(x_{h,t,a_t})]\\
    \leq & O\Bigg(\sum^H_{h=1} \sqrt{(1-\frac{\alpha}{2})T\left(\tilde{d}_h\log(1+\frac{(1-\frac{\alpha}{2})T}{\lambda}) +1 \right)\log\frac{1}{\delta}}\Bigg) + \\
    & O\Bigg(\sum^H_{h=1} \sqrt{(1-\frac{\alpha}{2})T\left(\tilde{d}_h\log(1+\frac{(1-\frac{\alpha}{2})T|\mathcal{A}|}{\lambda}) +1 \right)}(1+\sqrt{\lambda}+\lambda)S\Bigg) + \\
    & O\Bigg(\sum^H_{h=1} \sqrt{(1-\frac{\alpha}{2})T\left(\tilde{d}_h\log(1+\frac{(1-\frac{\alpha}{2})T}{\lambda}) +1 \right)\left(\tilde{d}\log(1+\frac{(1-\frac{\alpha}{2})T}{\lambda}) +1 \right)}\Bigg).
\end{align*}
The big $O$ notation hide the terms $\frac{\text{poly}(H,T,|\mathcal{A}|,\frac{1}{\lambda},\frac{1}{\lambda_0},\frac{1}{S},\log m)}{m^{\frac{1}{6}}}$ as they are bounded by the others, if we choose appropriate $m, J, \eta$.

\subsection{Supplemental Discussions on Existing Research on Contextual BwK}\label{app:supp_CBwK}
Existing research works \cite{BadanidiyuruKS18,AgrawalDL16,AgrawalD16,LiS22,HanZWXZ23,SlivkinsZSF24,GuoL24,ChenAYPWD24,GuoL25,LiZ23} replace the optimum $\textsf{opt}_t$ in episode $t$ with the benchmark $\textsf{UB}_t\geq \textsf{opt}_t$, which is the optimum of a linear program defined by the fluid relaxation of the problem. In our problem setting, in episode $t$ where the DM starts with $B_t$ units of resource, $\textsf{UB}_t$ is equal to the optimum of
\begin{equation}
        \begin{aligned}
             \max \quad & \sum^H_{h=1}\sum_{\theta\in \Theta} \sum_{a\in {\cal A}}\Pr_{\theta_h\sim \Lambda_h}(\theta_h = \theta)r(\theta, a)\rho(\theta, a)x_h(\theta, a) \\
            \text{s.t.} \quad & \sum^H_{h=1}\sum_{\theta\in \Theta} \sum_{a\in {\cal A}}\Pr_{\theta_h\sim \Lambda_h}(\theta_h = \theta)d(\theta, a)\rho(\theta, a)x_h(\theta, a) \leq B_t\quad,\\
            & \sum_{a\in {\cal A}}x_h(\theta, a) =1 \qquad\qquad \forall~h\in [H], \theta\in \Theta\\
           & x_h(\theta, a)\geq 0\qquad\qquad \qquad \forall~ h\in [H], \theta\in \Theta,  a \in \mathcal{A}.
        \end{aligned}
        \label{eq:UB-LP}
    \end{equation}
When there are $K\geq 1$ many resource types and the DM starts with $B_{t, k}$ units of resource $k\in [K]$, the single constraint in (\ref{eq:UB-LP}) shall be replaced by the set of constraints 
$$
\sum^H_{h=1}\sum_{\theta\in \Theta} \sum_{a\in {\cal A}}\Pr(\theta_h = \theta)d_d(\theta, a)\rho(\theta, a)x_h(\theta, a) \leq B_{t, k} \qquad \text{for all $k\in [K]$}
$$
in order for $\textsf{opt}_t\leq \textsf{UB}_t$ to hold true. The benchmark $\textsf{UB}_t$ above assumes a finite $\Theta$, and the optimization problem above has infinitely many decision variables $x_h(\theta, a)$'s when $|\Theta|=\infty.$

Importantly, existing results on $\text{Reg}(1) = o(H)$ requires that $\Lambda_1 = \ldots, \Lambda_H$ in (\ref{eq:UB-LP}) for their regret bounds of the form $\text{Reg}(1) = o(H)$ to hold. We show the more general fluid relaxation in the non-stationary $\Lambda_1, \ldots, \Lambda_H$ case for discussion sake.

\section{Proofs}
Our proofs use the Azuma-Hoeffding inequality, stated below.
\begin{proposition}[Azuma-Hoeffding Inequality]\label{prop:AH}
Let $X_1, \ldots, X_T$ be a martingale difference sequence adapted to the filtration ${\cal F}_0\subseteq {\cal F}_1\subseteq\ldots\subseteq {\cal F}_T$. That is, $X_t$ is ${\cal F}_t$-measurable, and $\mathbb{E}[X_t | {\cal F}_{t-1}] = 0$ with certainty. Furthermore, suppose that $|X_t|\leq b$ for all $t\in [T]$ almost surely. Then, for any $\delta\in (0, 1)$ it holds that
$$
\Pr\left( \sum^T_{t=1} X_t\leq b\sqrt{T\log(2/\delta)} \right) \geq 1-\delta.
$$ 
\end{proposition}

\subsection{Proof of Lemma \ref{lem:bound}}\label{app:pf_lemma_bound}
We focus on proving the upper bound $\hat{U}_{h, t}(b) - \hat{U}_{h, t}(b-d) \leq 2 r_\text{max} L$ in (\ref{eq:bound}), since the lower bound $0\leq \hat{U}_{h, t}(b) - \hat{U}_{h, t}(b-d) $ evidently holds for certainty, due to the fact that ${\cal A}(b, \theta)\subseteq {\cal A}(b', \theta)$ for any $b\leq b'\in [LH]$ and any $\theta\in \Theta$. We prove the upper bound $\hat{U}_{h, t}(b) - \hat{U}_{h, t}(b-d) \leq 2 r_\text{max} L$ in (\ref{eq:bound}) by a backward induction on $h$. For $h = H$, we observe that $\hat{V}_{H}(b, \theta) \in [0, r_{\text{max}}]$ for any $b\in \{0\}\cup [LH], \theta\in \Theta$, thus we also have $\hat{U}_H(b)\in [0, r_{\text{max}}]$ for any $b\in \{0\}\cup [LH], \theta\in \Theta$, hence (\ref{eq:bound}) holds for $h=H$. 

Assuming that (\ref{eq:bound}) holds for $h +1$, we prove that (\ref{eq:bound}) holds for $h$. In the case when $b<2L$,  we have 
$\hat{U}_{h, t}(b) - \hat{U}_{h, t}(b-d)\leq \hat{U}_{h, t}(2L)  \leq 2 r_\text{max} L$. The first inequality is by the monotonicity property $\hat{U}_{h, t}(b)\geq \hat{U}_{h, t}(b-d)$, the second inequality is because $\hat{U}_{h, t}(b)\leq b r_\text{max}$ by definition.

Next, in the case when $b\geq 2L$, we know that ${\cal A}(b, \theta) = {\cal A}(b-d, \theta)$ for any $\theta\in \Theta$, since $b-d\geq L$. For any $\theta\in \Theta$, we claim that
\begin{equation}\label{eq:bound_claim}
\hat{V}_{h, t}(b, \theta) - \hat{V}_{h, t}(b-d, \theta) \leq 2r_\text{max} L.
\end{equation}
Given (\ref{eq:bound_claim}), we can see that $\hat{U}_{h, t}(b) - \hat{U}_{h, t}(b-d) \leq 2 r_\text{max} L$ holds true by the following argument. By applying (\ref{eq:bound_claim}) with $\theta = \theta_{h, s}$ for every $s\in {\cal J}_t$ and averaging, we see that
\begin{equation}\label{eq:claim_ave}
\frac{1}{|{\cal J}_t|}\sum^{|{\cal J}_t|}_{s=1} \hat{V}_{h, t}(b, \theta_{h, s}) - \frac{1}{|{\cal J}_t|}\sum^{|{\cal J}_t|}_{s=1} \hat{V}_{h, t}(b-d, \theta_{h, s})\leq 2 r_\text{max} L.
\end{equation}
Thus, if $\frac{1}{|{\cal J}_t|}\sum^{|{\cal J}_t|}_{s=1} \hat{V}_{h, t}(b-d, \theta_{h, s}) \leq ((b-d)\wedge (H-h)r_{\text{max}}$, then $\hat{U}_{h, t}(b) - \hat{U}_{h, t}(b-d) \leq \text{left hand side of (\ref{eq:claim_ave})}\leq 2 r_\text{max}L$. Otherwise, we have $\hat{U}_{h, t}(b-d)  = ((b-d)\wedge (H-h))r_{\text{max}}$, and we still have
$$
\hat{U}_{h, t}(b) - \hat{U}_{h, t}(b-d)  \leq (b\wedge (H-h))r_{\text{max}} - ((b-d)\wedge (H-h))r_{\text{max}} \leq dr_\text{max} \leq L r_\text{max}. 
$$
Altogether, the induction argument is established.

To complete the proof, we prove (\ref{eq:bound_claim}). Now, for any $\bar{b}\in \{0\}\cup [LH]$, we denote $\hat{a}_{\bar{b}}\in {\cal A}(\bar{b}, \theta)$ as an optimal solution to the maximization problem on the right hand side:
\begin{align*}
    \hat{V}_{h, t}(\bar{b}, \theta) &= \max_{a\in {\cal A}(\bar{b},\theta)} \left\{\text{UCB}_t(\theta, a;\delta_t) r(\theta, a) \right.\nonumber\\
	&\left. + \text{LCB}_t(\theta, a;\delta_t) \hat{U}_{h+1,t}(\bar{b}- d(\theta, a)) + \left[1-\text{LCB}_t(\theta, a;\delta_t)\right] \hat{U}_{h+1,t}(\bar{b})\right\}.
\end{align*}
We assert that the following (in)equalities hold with certainty:
\begin{align}
    &\hat{V}_{h, t}(b, \theta) - \hat{V}_{h, t}(b-d, \theta)\nonumber\\
=& \text{UCB}_t(\theta, \hat{a}_b;\delta_t) r(\theta, \hat{a}_b)  + \text{LCB}_t(\theta, \hat{a}_b;\delta_t) \hat{U}_{h+1,t}(b- d(\theta, \hat{a}_b)) + \left[1-\text{LCB}_t(\theta, \hat{a}_b;\delta_t)\right] \hat{U}_{h+1,t}(b)\nonumber\\
-&  \left\{ \text{UCB}_t(\theta, \hat{a}_{b-d};\delta_t) r(\theta, \hat{a}_{b-d})  + \text{LCB}_t(\theta, \hat{a}_{b-d};\delta_t) \hat{U}_{h+1,t}((b- d )- d(\theta, \hat{a}_{b-d}))\right.\nonumber\\
&\qquad \left.+ \left[1-\text{LCB}_t(\theta, \hat{a}_{b-d};\delta_t)\right] \hat{U}_{h+1,t}(b-d)\right\}\nonumber\\
\leq& \text{UCB}_t(\theta, \hat{a}_b;\delta_t) r(\theta, \hat{a}_b)  + \text{LCB}_t(\theta, \hat{a}_b;\delta_t) \hat{U}_{h+1,t}(b- d(\theta, \hat{a}_b)) + \left[1-\text{LCB}_t(\theta, \hat{a}_b;\delta_t)\right] \hat{U}_{h+1,t}(b)\nonumber\\
-&  \left\{ \text{UCB}_t(\theta, \hat{a}_{b};\delta_t) r(\theta, \hat{a}_{b})  + \text{LCB}_t(\theta, \hat{a}_{b};\delta_t) \hat{U}_{h+1,t}((b- d )- d(\theta, \hat{a}_{b}))\right.\nonumber\\
&\qquad \left.+ \left[1-\text{LCB}_t(\theta, \hat{a}_{b};\delta_t)\right] \hat{U}_{h+1,t}(b-d)\right\}\label{eq:by_large}\\
=& \text{LCB}_t(\theta, \hat{a}_b;\delta_t) \left[ \hat{U}_{h+1,t}(b- d(\theta, \hat{a}_b)) - \hat{U}_{h+1,t}((b -d) - d(\theta, \hat{a}_b)) \right] \nonumber\\
&\qquad + \left[1-\text{LCB}_t(\theta, \hat{a}_b;\delta_t)\right] \left[ \hat{U}_{h+1,t}(b) - \hat{U}_{h+1,t}(b-d)\right] \nonumber\\
\leq &\text{LCB}_t(\theta, \hat{a}_b;\delta_t) \cdot (2 r_\text{max}L) + [1-\text{LCB}_t(\theta, \hat{a}_b;\delta_t) ] \cdot (2 r_\text{max}L) = 2 r_\text{max}L\label{eq:by_induction}.
\end{align}
Step (\ref{eq:by_large}) is because ${\cal A}(b, \theta) ={\cal A}(b-d, \theta)$. The inequality in step (\ref{eq:by_induction}) is by the induction hypothesis of (\ref{eq:bound}) at time step $h+1$, and the fact that $\text{LCB}_t(\theta, \hat{a}_b;\delta_t) \in [0,1]$. Altogether, (\ref{eq:bound_claim}) is shown. 

Lastly, the argument above applies to any realization of $\text{LCB}_t, \text{UCB}_t$. By specializing $\text{LCB}_t(\theta, a ; \delta) = \text{UCB}_t(\theta, a ; \delta) = \rho(\theta, a)$, we have 
\begin{equation}\label{eq:bound_last}
\hat{U}_{h, t}(b) = \frac{1}{|{\cal J}_t|}\sum_{s\in {\cal J}_t} V_h(b, \theta_{h,s}) \wedge (b\wedge(H-h+1))r_\text{max} = \frac{1}{|{\cal J}_t|}\sum_{s\in {\cal J}_t} V_h(b, \theta_{h,s }),
\end{equation}
where the last equality in (\ref{eq:bound_last}) is because $V_h(b, \theta) \leq (b\wedge(H-h+1))r_\text{max}$, which is true since Assumption \ref{ass:nonnull} says that if the DM earns a reward of $r(A_{h,t}, \theta_{h,t}) > 0$ in a time step, then the DM must consumes at least 1 unit of the resource. Taking expectation in (\ref{eq:bound_last}) gives $\mathbb{E}[\hat{U}_{h,t}(b)] = U_h(b)$, hence the analogous inequalities on $U_h$ are also shown, and the proof is complete.

\subsection{Proof of Lemma \ref{lemma:optimism}}\label{app:pf_lemma_optimism}
Before we start the proof, we remind that for proving Theorem \ref{thm:main} it suffices to put $M = 0$, while $M>0$ is only needed for establishing Corollary \ref{cor:unlabeled}. Recall the notation $U_h(b) = \mathbb{E}_{\theta_h\sim \Lambda_h} \left[V_h(b, \theta_h)\right]$. By an abuse of notation, we set $U_{H+1}(b) = 0$ for all $b\in \{0\}\cup [LH]$, similarly to how we have set $\hat{U}_{H+1, t}(b) = 0$ for all $b\in \{0\}\cup [LH]$ when we invoke Algorithm \ref{alg:UCB}. Consider the event
\begin{equation*}
    {\cal E}_{h, t} = \left\{U_{h}(b) \leq \hat{U}_{h, t}(b) +  r_\text{max} (2L+1) (H-h+1)\sqrt{\frac{\log(2HB/\delta')}{|{\cal J}_t|+M}}\text{ for all $b\in \{0\}\cup [LH]$}\right\}
\end{equation*}
for each $h\in [H+1]$, and recall that 
\begin{equation*}
    {\cal G}_t = \left\{\rho(\theta, a)\in [\text{LCB}_t(\theta,a;\delta_t), \text{UCB}_t(\theta, a;\delta_t)] \text{ for all $\theta\in \Theta, a\in {\cal A}$}\right\}.
\end{equation*}
We prove the inequality 
\begin{equation}\label{eq:optimisim_induction}
    \Pr\left({\cal E}_{h,t}\cap {\cal G}_t\right)\geq 1-\delta_t - \frac{(H-h+1)\delta'}{H}
\end{equation}
by a backward induction on $h = H+1, \ldots, 1$. Once the induction proof is completed, the Lemma is established since if the event ${E}_{1, t}$ holds then the event ${\cal F}_t$ also holds. Firstly, for the basic induction hypothesis when $h = H+1$, it is clear that $\Pr({\cal E}_{H+1,t})=1$ since we have defined $U_{H+1}(b) = \hat{U}_{H+1, t}(b) = 0$ for all $b\in \{0\}\cup [LH]$. Thus, $\Pr({\cal E}_{H+1,t}\cap {\cal G}_t)=\Pr({\cal G}_t)\geq 1-\delta_t$ by the definition of a CB oracle (see Assumption \ref{ass:conf}). 

In what follows, we show that 
\begin{equation}\label{eq:optimism_crucial}
\Pr({\cal E}_{h, t} | {\cal E}_{h+1, t}\cap  {\cal G}_t) \geq 1 - \frac{\delta'}{H}, 
\end{equation}
which establishes the induction argument, since multiplying (\ref{eq:optimisim_induction}) with $h$ replaced by $h+1$ and (\ref{eq:optimism_crucial}) gives
$$
\Pr({\cal E}_{h, t} \cap {\cal E}_{h+1, t}\cap  {\cal G}_t) \geq \left[1 - \frac{\delta'}{H}\right]\cdot \left[ 1-\delta_t - \frac{(H-h)\delta'}{H}\right] \geq  1-\delta_t - \frac{(H-h+1)\delta'}{H}.
$$
For the remaining part of the proof, we focus on proving (\ref{eq:optimism_crucial}). We now claim that, for each $s\in {\cal J}_t$ (so that $\theta_{h, s}\in {\cal S}_t$) and for each $b\in \{0\}\cup [LH]$, we have
$$
\Pr\left( \hat{V}_{h, t}(b, \theta_{h,s}) \geq V_h(b, \theta_{h,s}) - r_\text{max}(2L+1) (H-h)\sqrt{\frac{\log(2HB/\delta')}{|{\cal J}_t|+M}} \mid {\cal E}_{h+1, t}\cap {\cal G}_t \right) = 1.
$$
Recall that $\pi^*_h(b, \theta_{h,s})$ an optimal action to take at time step $h$, when the DM has $b$ units of inventory left and he sees a request of type $\theta_{h,s }$ arrives (see Section \ref{sec:bellman}). The above equality can be established by following inequalities, which all hold with certainty:
\begin{align}
    &\hat{V}_{h, t}(b, \theta_{h,s})\nonumber\\
    \geq &  \text{UCB}_t(\theta_{h,s}, \pi^*_h(b, \theta_{h,s});\delta_t) r(\theta_{h,s},  \pi^*_h(b, \theta_{h,s})) \nonumber\\
    &\qquad + \text{LCB}_t(\theta_{h,s},  \pi^*_h(b, \theta_{h,s});\delta_t) \hat{U}_{h+1,t}(b- d(\theta_{h,s},  \pi^*_h(b, \theta_{h,s})))\nonumber\\
    &\qquad\qquad + \left[1-\text{LCB}_t(\theta_{h,s},  \pi^*_h(b, \theta_{h,s});\delta_t)\right] \hat{U}_{h+1,t}(b)\label{eq:by_optimality}\\
    \geq  & \rho(\theta_{h,s}, \pi^*_h(b, \theta_{h,s}))r(\theta_{h,s},  \pi^*_h(b, \theta_{h,s})) + \rho(\theta_{h,s}, \pi^*_h(b, \theta_{h,s})) \hat{U}_{h+1,t}(b- d(\theta_{h,s},  \pi^*_h(b, \theta_{h,s})))\nonumber\\
    &\qquad + \left[1-\rho(\theta_{h,s}, \pi^*_h(b, \theta_{h,s}))\right] \hat{U}_{h+1,t}(b)
    \label{eq:by_optimism}\\
    \geq & \rho(\theta_{h,s}, \pi^*_h(b, \theta_{h,s}))r(\theta_{h,s},  \pi^*_h(b, \theta_{h,s})) + \rho(\theta_{h,s}, \pi^*_h(b, \theta_{h,s})) U_{h+1}(b- d(\theta_{h,s},  \pi^*_h(b, \theta_{h,s})))\nonumber\\
    &\qquad + \left[1-\rho(\theta_{h,s}, \pi^*_h(b, \theta_{h,s}))\right] U_{h+1}(b)- r_\text{max}(2L+1) (H-h)\sqrt{\frac{\log(2HB/\delta')}{|{\cal J}_t|+M}}\label{eq:by_optimism_induction}\\
    = & V_h(b, \theta_{h,s}) - r_\text{max}(2L+1) (H-h)\sqrt{\frac{\log(2HB/\delta')}{|{\cal J}_t|+M}}.\label{eq:optimism_by_defn_V}
\end{align}
Step (\ref{eq:by_optimality}) is because we have $\pi^*_h(b, \theta_{h,s}))\in {\cal A}(b, \theta_{h,s})$, meaning that $\pi^*_h(b, \theta_{h,s}))$ is a feasible solution to the maximization problem in (\ref{eq:hatVt}) with $\theta$ set to be $\theta_{h,s}$. Step (\ref{eq:by_optimism}) is because we condition on the event ${\cal G}_t$, which means that $\rho(\theta_{h,s}, \pi^*_h(b, \theta_{h,s}))\in [\text{LCB}_t(\theta_{h,s}, \pi^*_h(b, \theta_{h,s})), \text{UCB}_t(\theta_{h,s}, \pi^*_h(b, \theta_{h,s}))]$, and by the monotonicity property that $ \hat{U}_{h+1,t}(b- d(\theta_{h,s},  \pi^*_h(b, \theta_{h,s})))\leq  \hat{U}_{h+1,t}(b)$. Step (\ref{eq:by_optimism_induction}) is because we condition on the event ${\cal E}_{h+1, t}$. Lastly, step (\ref{eq:optimism_by_defn_V}) is by the definition of $V_h$.

Lastly, we prove (\ref{eq:optimism_crucial}) by the following:
\begin{align}
    &\Pr({\cal E}_{h, t} | {\cal E}_{h+1, t}\cap  {\cal G}_t) \nonumber\\
    \geq & \Pr\left( {\cal E}_{h,t}\cap \left\{ \hat{V}_{h, t}(b, \theta_{h,s}) \geq V_h(b, \theta_{h,s}) -  r_\text{max}(2L+1) (H-h)\sqrt{\frac{\log(2HB/\delta')}{|{\cal J}_t|+M}}  ~\forall b, s \right\} \mid {\cal E}_{h+1, t}\cap  {\cal G}_t \right)\label{eq:optimism_short}\\
    \geq & \Pr\left( \frac{1}{|{\cal J}_t|+M} \sum_{s\in {\cal J}_t\cup \{-1, \ldots, -M\}}   V_h(b, \theta_{h, s}) \geq U_h(b)- r_\text{max}(2L+1) \sqrt{\frac{\log(2HB/\delta')}{|{\cal J}_t|+M}}\right.\nonumber\\
    &\qquad \qquad \text{ for all $b\in \{0\}\cup [LH]$}\mid {\cal E}_{h+1, t}\cap  {\cal G}_t \bigg)\nonumber\\
    = & \Pr\left( \frac{1}{|{\cal J}_t|+M} \sum_{s\in {\cal J}_t\cup \{-1, \ldots, -M\}}   V_h(b, \theta_{h, s}) \geq U_h(b)-r_\text{max}(2L+1) \sqrt{\frac{\log(2HB/\delta')}{|{\cal J}_t|+M}} \text{ for all $b\in \{0\}\cup [LH]$} \right)\label{eq:independence}\\
    \geq & 1 - LH\cdot \frac{\delta'}{LH^2}\label{eq:optimisim_by_AH}.
\end{align}
In Step (\ref{eq:optimism_short}), $\forall b, s$ refers to for all $b\in \{0\}\cup [LH]$ and $s\in {\cal J}_t$. Step (\ref{eq:independence}) is because the random features in $\{\theta_{h, s}\}_{s\in {\cal J}_t\cup \{-1, \ldots, -M\}}$ are independent of the events ${\cal E}_{h+1, t}, {\cal G}_t$. The independence from ${\cal G}_t$ is because $\{\theta_{h, s}\}_{s\in {\cal J}_t\cup \{-1, \ldots, -M\}}$ are independent of $\text{UCB}_t, \text{LCB}_t$, which are constructed using the dataset ${\cal D}_t$ that is disjoint from $\{\theta_{h, s}\}_{s\in {\cal J}_t}$. This illustrates the reason behind Lines \ref{alg:begin_split}--\ref{alg:end_split} in Algorithm \ref{alg:online}, where we use the observed data to either estimate $\rho$ or to estimate $\Lambda_1, \ldots, \Lambda_H$, but not both. The independence from ${\cal E}_{h+1, t}$ is because the random variables involved in ${\cal E}_{h+1, t}$ are $\text{UCB}_t, \text{LCB}_t$ and $\{\theta_{\tau, s}\}_{s\in {\cal J}_t\cup \{-1, \ldots, -M\}, h+1\leq \tau\leq H}$, and the latter is independent of $\{\theta_{h, s}\}_{s\in {\cal J}_t\cup \{-1, \ldots, -M\}}$ by the independence among requests' features. 

Step (\ref{eq:optimisim_by_AH}) is by the application of the Azuma-Hoeffding inequality (see Proposition \ref{prop:AH}), where we observed that $\mathbb{E}_{\theta_{h, s}\sim\Lambda_h}[V_h(b, \theta_{h, s})] = U_h(b)$, and that $|V_h(b, \theta_{h, s}) - U_h(b)| \leq r_\text{max}(2L+1)$ by Corollary \ref{cor:bound_V}. The bound (\ref{eq:optimisim_by_AH}) finally follows by a union bound over $b\in [B]$. Altogether, (\ref{eq:optimism_crucial}) is proved, and the Lemma is also proved.

\subsection{Proof of Theorem \ref{thm:regt}}\label{app:pf_thm_regt}
Before we start the proof, we recall the event ${\cal F}_{t, \delta'}$ defined in Lemma \ref{lemma:optimism}, and the event ${\cal G}_{t, \delta_t}$ defined in (\ref{eq:event_cb}). The event $\tilde{\cal E}_{h,t, \delta'}$ is defined in (\ref{eq:event_E_tilde}) in this Appendix section. In addition, $(\star_t) = (\diamondsuit_t)+(\dagger_t) + (\ddagger_t)$. The quantities $(\diamondsuit_t),(\dagger_t), (\ddagger_t)$ are defined in the forthcoming (\ref{eq:diamond}, \ref{eq:dagger}, \ref{eq:ddagger}) respectively, and the inequality $\Pr(\sum^T_{t=1}(\star_t) \leq 2r_\text{max}(2L+1) \sqrt{HT\log(2/\delta'')}) \geq 1-3\delta''$ for any $\delta''\in (0,1)$ is justified by a union bound on the forthcoming (\ref{eq:diamond_by_AH}, \ref{eq:dagger_bound},\ref{eq:ddagger_bound}).

We embark on the proof by the following decomposition.
\begin{align}
    &\textsf{opt}_t - \sum^H_{h=1}R_{h, t} \nonumber\\
=& \underbrace{\hat{U}_{1, t}(b_{1, t}) - \sum^H_{h=1} r(\theta_{h, t}, A_{h, t})\cdot \text{UCB}_t(\theta_{h, t}, A_{h, t})}_{(\spadesuit_t)}  +  \underbrace{\textsf{opt}_t -  \hat{U}_{1, t}(B_t)}_{(\clubsuit_t)} \nonumber\\
+&\underbrace{\sum^H_{h=1} r(\theta_{h, t}, A_{h, t})\cdot \left[\text{UCB}_t(\theta_{h, t}, A_{h, t}) - \rho(\theta_{h, t}, A_{h, t})\right]}_{(\heartsuit_t)} +  \underbrace{\sum^H_{h=1} r(\theta_{h, t}, A_{h, t})\cdot \rho(\theta_{h, t}, A_{h, t}) - \sum^H_{h=1}R_{h, t}}_{(\diamondsuit_t)}\label{eq:diamond},
\end{align}
and in the above we recall that $B_t = b_{1, t}$. The term $(\spadesuit_t)$ is the main term to analyze, while the remaining terms are bounded from above as:
\begin{align}
    &(\clubsuit_t) \leq r_\text{max}(2L+1)H \sqrt{\frac{\log(2LH^2/\delta')}{|{\cal J}_t|+M}} + r_\text{max} B_t\cdot \mathbf{1}(\neg {\cal F}_{t, \delta'}) \text{ almost surely},\label{eq:by_lemma_optimisim}\\
    &(\heartsuit_t) \leq r_\text{max}\sum^H_{h=1} [\text{UCB}_t(\theta_{h, t},A_{h,t}) - \text{LCB}_t(\theta_{h, t},A_{h,t})] + r_\text{max} H \mathbf{1}(\neg {\cal G}_{t, \delta_t})\text{ almost surely},\label{eq:by_cb_oracle}\\
    &\Pr\left(\sum^T_{t=1} (\diamondsuit_t) \leq r_\text{max} \sqrt{HT\log(2/\delta'')}\right) \geq 1-\delta'' \text{ for any $\delta''\in (0, 1)$}. \label{eq:diamond_by_AH}
\end{align}
The bound (\ref{eq:by_lemma_optimisim}) is by Lemma \ref{lemma:optimism}, the bound (\ref{eq:by_cb_oracle}) is by the notion of CB oracle (see Assumption \ref{ass:conf}). The bound (\ref{eq:diamond_by_AH}) is by the Azuma-Hoeffding inequality (see Proposition \ref{prop:AH}), and we note that $\mathbb{E}_{Y_{h,t}\sim\rho(\theta_{h,t}, A_{h,t})}[r(\theta_{h, t}, A_{h, t})\cdot \rho(\theta_{h, t}, A_{h, t}) - R_{h, t} ~|~ \theta_{h,t}, A_{h,t}] = 0$ and $|r(\theta_{h, t}, A_{h, t})\cdot \rho(\theta_{h, t}, A_{h, t}) - R_{h, t}|\leq r_\text{max}$ with certainty.

We demonstrate the following upper bound to $(\spadesuit_t)$, which holds almost surely:
\begin{align}
    &\hat{U}_{1, t}(b_{1, t}) - \sum^H_{h = 1} r(\theta_{h, t}, A_{h, t})\cdot \text{UCB}_t(\theta_{h, t}, A_{h, t})\nonumber\\
    \leq &r_\text{max} (2L+1) H\sqrt{\frac{\log(2HB/\delta')}{|{\cal J}_t|+M}} +  r_\text{max} (2L+1) \sum^H_{h=1}\mathbf{1}(\neg \tilde{\cal E}_{h,t, \delta'})+ \sum^H_{h=1}(\ddagger_{h,t}) \nonumber\\
    &+2 r_\text{max} L \sum^H_{h=1} \left[\text{UCB}_t(\theta_{h,t}, A_{h,t};\delta_t) - \text{LCB}_t(\theta_{h,t}, A_{h,t};\delta_t)\right] + 2 r_\text{max} L H \cdot \mathbf{1}(\neg {\cal G}_{t, \delta_t}) + \sum^H_{h=1}(\dagger_{h,t}), \label{eq:spade_bound} 
\end{align}
where the event $\tilde{\cal E}_{h,t, \delta'}$ is defined in the forthcoming (\ref{eq:event_E_tilde}), and the error terms $(\dagger_{h,t}, \ddagger_{h,t})$ are defined in the forthcoming (\ref{eq:dagger}, \ref{eq:ddagger}) respectively. To prove (\ref{eq:spade_bound}), consider a fixed time step $(h, t)$ where $h\in [H]$:
\begin{align}
    &\hat{U}_{h, t}(b_{h, t}) - \sum^H_{\tau = h} r(\theta_{\tau, t}, A_{\tau, t})\cdot \text{UCB}_t(\theta_{\tau, t}, A_{\tau, t})\nonumber\\
    \leq & \frac{1}{|{\cal J}_t|+M}\sum_{s\in {\cal J}_t\cup \{-1, \ldots, -M\}}\hat{V}_{h,t}(b_{h, t}, \theta_{h,s}) - r(\theta_{h, t}, A_{h, t})\cdot \text{UCB}_t(\theta_{h, t}, A_{h, t}) \nonumber\\
    &\qquad - \sum^H_{\tau=h+1} r(\theta_{\tau, t}, A_{\tau, t})\cdot \text{UCB}_t(\theta_{\tau, t}, A_{\tau, t})\label{eq:by_def_hatU}.
\end{align}
Step (\ref{eq:by_def_hatU}) is by the definition of $\hat{U}_{\tau, t}$ in Line \ref{alg:UCB_Uhat} in Algorithm \ref{alg:UCB}. We consider the following decomposition:
\begin{align}
    &\frac{1}{|{\cal J}_t|+M}\sum_{s\in {\cal J}_t\cup \{-1, \ldots, -M\}}\hat{V}_{h,t}(b_{h, t}, \theta_{h,s}) - r(\theta_{h, t}, A_{h, t})\cdot \text{UCB}_t(\theta_{h, t}, A_{h, t})\nonumber\\
   = & \frac{1}{|{\cal J}_t|+M}\sum_{s\in {\cal J}_t\cup \{-1, \ldots, -M\}}\hat{V}_{h,t}(b_{h, t}, \theta_{h,s}) - \hat{V}_{h,t}(b_{h, t}, \theta_{h,t}) \label{eq:term1}\\
    + & \text{LCB}_t(\theta_{h,t}, A_{h,t};\delta_t) \hat{U}_{h+1,t}(b_{h,t}- d(\theta_{h,t}, A_{h,t})) + \left[1-\text{LCB}_t(\theta_{h,t}, A_{h,t};\delta_t)\right] \hat{U}_{h+1,t}(b_{h,t}).\label{eq:mimic}
\end{align}
The equality (\ref{eq:mimic}) is by the definition of $\hat{V}_{h,t}$ in (\ref{eq:hatVt}) and the selection rule of $A_{h,t}$ in Line \ref{alg:online_Aht} in Algorithm \ref{alg:online}, which mimics the selection rule by $\hat{V}_{h,t}$. 

We analyze (\ref{eq:term1}, \ref{eq:mimic}) in what follows. In (\ref{eq:term1}), it is important to observe that, for each $\theta_{h, s}$ where $s\in {\cal J}_s\cup \{-1, \ldots, -M\}$, the random variable $\theta_{h,s}$ is independent of $\hat{V}_{h, t}$, since the random function $\hat{V}_{h, t}$ is constructed using the datasets ${\cal D}_t$ and $\{\theta_{\tau, s}\}_{s\in {\cal J}_t\cup \{-1, \ldots, -M\}, h+1\leq \tau\leq H}$. The dataset ${\cal D}_t$ is disjoint from ${\cal S}_t$ that contains $\{\theta_{h, s}\}_{s\in {\cal J}_t\cup \{-1, \ldots, -M\}}$ by construction, and the former is used to construct $\text{UCB}_t, \text{LCB}_t$ via the confidence bound oracle. The independence from $\{\theta_{\tau, s}\}_{s\in {\cal J}_t\cup \{-1, \ldots, -M\}, h+1\leq \tau\leq H}$ is by the fact that the requests' features in different time steps are jointly independent. Next, we observe that $\theta_{h, s}$ is \emph{not independent of} $b_{h, s}$ in general, since $b_{h, s}$ is dependent on the online allocation policy at time $t$, and the policy is dependent on  ${\cal D}_t, {\cal S}_t$. 

To aid our analysis, for each $b\in \{0\}\cup [LH]$, we consider the conditional expectation $\mathbb{E}_{\theta_h\sim \Lambda_h}[\hat{V}_{h,t}(b, \theta_h) ~|~ \hat{V}_{h, t}]$, which is conditioning on the realization of the random function $\hat{V}_{h, t}$ and only taking expectation over the randomness of $\theta_h\sim \Lambda_h$. Consider the event 
\begin{align}
    &\tilde{\cal E}_{h,t, \delta'} =\left\{ \frac{1}{|{\cal J}_t|+M}\sum_{s\in {\cal J}_t\cup \{-1, \ldots, -M\}}\hat{V}_{h,t}(b, \theta_{h,s})  -  \mathbb{E}_{\theta_h\sim \Lambda_h}[\hat{V}_{h,t}(b, \theta_h) ~|~ \hat{V}_{h, t}] \right.\nonumber\\
    &\qquad\qquad\qquad \qquad  \left. \leq r_\text{max} (2L+1) \sqrt{\frac{\log(2HB/\delta')}{|{\cal J}_t|+M}}\text{ for all $b\in \{0\}\cup [LH]$}\right\}.\label{eq:event_E_tilde}
\end{align}
By the Azuma-Hoeffding inequality and Corollary \ref{cor:bound_V}, we know that $\Pr({\cal E}_{h, t, \delta'}) \geq 1 - \delta' / H$. Now, with certainty we have
\begin{align}
    (\ref{eq:term1}) &= \frac{1}{|{\cal J}_t|+M}\sum_{s\in {\cal J}_t\cup \{-1, \ldots, -M\}}\hat{V}_{h,t}(b_{h, t}, \theta_{h,s}) - \mathbb{E}_{\theta_h\sim \Lambda_h}[\hat{V}_{h,t}(b_{h, t}, \theta_h) ~|~ \hat{V}_{h, t},b_{h,t}]\label{eq:term1_contt}\\
    & \qquad \qquad \qquad + \underbrace{\mathbb{E}_{\theta_h\sim \Lambda_h}[\hat{V}_{h,t}(b_{h,t}, \theta_h) ~|~ \hat{V}_{h, t},b_{h,t}]  - \hat{V}_{h,t}(b_{h, t}, \theta_{h,t})}_{=(\dagger_{h,t})} \label{eq:dagger}\\
    & \leq r_\text{max} (2L+1) \sqrt{\frac{\log(2HB/\delta')}{|{\cal J}_t|}} + r_\text{max} (2L+1) \cdot \mathbf{1}(\neg \tilde{\cal E}_{h,t, \delta'}) + (\dagger_{h,t}). \label{eq:term1_cont}
\end{align}
Step (\ref{eq:term1_cont}) is because if event ${\cal E}_{h, t, \delta'}$ holds, then we have  (\ref{eq:term1_contt}) $\leq r_\text{max} (2L+1) \sqrt{\frac{\log(2LH^2/\delta')}{|{\cal J}_t|+M}}$. Indeed, the event ${\cal E}_{h,t, \delta'}$ asserts that the upper bound on the right hand side holds for any realization of $b_{h,t}.$ Notice that $\mathbb{E}_{\theta_{h,t}\sim \Lambda_h}[(\dagger_{h,t}) |\hat{V}_{h,t}, b_{h,t} ] = 0$, and we have $|(\dagger)_{h,t}|\leq r_\text{max}(2L+1)$ with certainty by Corollary \ref{cor:bound_V}. By the Azuma-Hoeffding inequality, we have
\begin{equation}\label{eq:dagger_bound}
\Pr\left(\sum^T_{t=1} \sum^H_{h=1} (\dagger_{h,t})\leq r_\text{max}(2L+1)\sqrt{HT\log\frac{2}{\delta''}}\right)\geq 1-\delta'' \text{ for any $\delta''\in (0,1)$}.
\end{equation}

Next, we analyze (\ref{eq:mimic}) by the following decomposition:
\begin{align}
&(\ref{eq:mimic})\nonumber\\
     =& \text{LCB}_t(\theta_{h,t}, A_{h,t};\delta_t) \hat{U}_{h+1,t}(b_{h,t}- d(\theta_{h,t}, A_{h,t})) + \left[1-\text{LCB}_t(\theta_{h,t}, A_{h,t};\delta_t)\right] \hat{U}_{h+1,t}(b_{h,t})\nonumber\\
     = & \text{LCB}_t(\theta_{h,t}, A_{h,t};\delta_t) \hat{U}_{h+1,t}(b_{h,t}- d(\theta_{h,t}, A_{h,t})) + \left[1-\text{LCB}_t(\theta_{h,t}, A_{h,t};\delta_t)\right] \hat{U}_{h+1,t}(b_{h,t})\nonumber\\
     & - \rho(\theta_{h,t}, A_{h,t}) \hat{U}_{h+1,t}(b_{h,t}- d(\theta_{h,t}, A_{h,t})) - \left[1-\rho(\theta_{h,t}, A_{h,t})\right] \hat{U}_{h+1,t}(b_{h,t})\label{eq:diff_1}\\
      +&\underbrace{\rho(\theta_{h,t}, A_{h,t}) \hat{U}_{h+1,t}(b_{h,t}- d(\theta_{h,t}, A_{h,t})) + \left[1-\rho(\theta_{h,t}, A_{h,t})\right] \hat{U}_{h+1,t}(b_{h,t}) - \hat{U}_{h+1,t}(b_{h,t} -  Y_{h, t}d(\theta_{h,t}, A_{h,t})) }_{=(\ddagger_{h,t})}\label{eq:ddagger}\\
      &\qquad\qquad\qquad\qquad  +\hat{U}_{h+1,t}(b_{h+1,t} ).\nonumber
\end{align}
Now, we analyze each term as follows. For 
\begin{align}
    (\ref{eq:diff_1})  = & \left(\rho(\theta_{h,t}, A_{h,t}) - \text{LCB}_t(\theta_{h,t}, A_{h,t};\delta_t)\right) \cdot \left( \hat{U}_{h+1,t}(b_{h,t}- d(\theta_{h,t}, A_{h,t}))- \hat{U}_{h+1,t}(b_{h,t})   \right) \nonumber\\
     \leq & 2 r_\text{max} L \cdot \left(\text{UCB}_t(\theta_{h,t}, A_{h,t};\delta_t) - \text{LCB}_t(\theta_{h,t}, A_{h,t};\delta_t)\right) + 2 r_\text{max} L \cdot \mathbf{1}(\neg {\cal G}_{t, \delta_t}). \label{eq:diff_1_reason}
\end{align}
Step (\ref{eq:diff_1_reason}) is by Lemma \ref{lem:bound} and the definition of the good event ${\cal G}_{t, \delta_t}$. Next, we observe that $\mathbb{E}_{Y_{h,t}\sim \rho( \theta_{h,t}, A_{h,t})  }[(\ddagger_{h, t}) ~|~ \hat{U}_{h+1, t}, \theta_{h,t}, A_{h,t} ] = 0$, and that $|(\ddagger_{h,t})|\leq 2r_\text{max} L$ with certainty by Lemma \ref{lem:bound}. Thus the application of the Azuma-Hoeffiding inequality gives
\begin{equation}\label{eq:ddagger_bound}
    \Pr\left( \sum^T_{t=1}\sum^H_{h=1}(\ddagger_{h,t})\leq 2r_\text{max} L \sqrt{HT\log\frac{2}{\delta''}}   \right)\geq 1-\delta'' \text{ for any $\delta''\in (0,1)$.}
\end{equation}
Applying our analyses on (\ref{eq:term1}, \ref{eq:mimic}) onto (\ref{eq:by_def_hatU}) gives
\begin{align}
    &\hat{U}_{h, t}(b_{h, t}) - \sum^H_{\tau = h} r(\theta_{\tau, t}, A_{\tau, t})\cdot \text{UCB}_t(\theta_{\tau, t}, A_{\tau, t})\nonumber\\
    \leq & \hat{U}_{h+1, t}(b_{h+1, t}) - \sum^H_{\tau = h+1} r(\theta_{\tau, t}, A_{\tau, t})\cdot \text{UCB}_t(\theta_{\tau, t}, A_{\tau, t}) \nonumber\\
    &\qquad + r_\text{max} (2L+1) \sqrt{\frac{\log(2HB/\delta')}{|{\cal J}_t|+M}} + r_\text{max} (2L+1) \cdot \mathbf{1}(\neg \tilde{\cal E}_{h,t, \delta'}) + (\dagger_{h,t}) \nonumber\\
    &+ 2 r_\text{max} L \cdot \left(\text{UCB}_t(\theta_{h,t}, A_{h,t};\delta_t) - \text{LCB}_t(\theta_{h,t}, A_{h,t};\delta_t)\right) + 2 r_\text{max} L \cdot \mathbf{1}(\neg {\cal G}_{t, \delta_t}) + (\ddagger_{h,t}) \label{eq:spade_induction}. 
\end{align}
Finally, applying (\ref{eq:spade_induction}) on $h = 1, \ldots, H$ gives (\ref{eq:spade_bound}). By establishing the required bound (\ref{eq:spade_bound}) on $(\spadesuit_t)$, the Theorem is proved. 

\subsection{Proof of Lemma \ref{lem:final_AH}}\label{app:pf_final_AH}
Consider the set ${\cal I}_{T+1} = \{t\in [T] : t+1\in {\cal J}_{T+1}\}$. Notice that $|{\cal I}_{T+1}| = |{\cal J}_{T+1}| - 1$. By the definition of ${\cal J}_{T+1}$, we know that ${\cal I}_{T+1}\subset [T]\setminus {\cal J}_{T+1}$. For each $t\in {\cal I}_{T+1}$, the following statements in this paragraph hold true. We have $\text{LCB}_t=\text{LCB}_{t+1}, \text{UCB}_t=\text{UCB}_{t+1}$, $\hat{U}_{h, t} = \hat{U}_{h, t+1}$, by the design of Line \ref{alg:Jt_static} in Algorithm \ref{alg:online}. Consequently, the two random vectors $((\theta_{h, t}, A_{h,t}))^H_{h=1}, ((\theta_{h, t+1}, A_{h,t+1}))^H_{h=1}$ are identically distributed. Combining the two previous observations, the two random variables $\sum^H_{h=1} [\text{UCB}_t(\theta_{h, t},A_{h,t}) - \text{LCB}_t(\theta_{h, t},A_{h,t})]$, $\sum^H_{h=1} [\text{UCB}_{t+1}(\theta_{h, t+1},A_{h,t+1}) - \text{LCB}_{t+1}(\theta_{h, t+1},A_{h,t+1})]$ are identically distributed. 

Next, we set up some notation to apply the Azuma-Hoeffding inequality. Denote ${\cal I}_{T+1} = \{t_1, \ldots, t_m\}$, where $t_1< t_2 < \ldots <t_m$. Define 
\begin{align*}
X_n &= -\sum^H_{h=1} [\text{UCB}_{t_n}(\theta_{h, t_n},A_{h,t_n}) - \text{LCB}_{t_n}(\theta_{h, t_n},A_{h,t_n})]\nonumber\\
&\qquad \qquad + \sum^H_{h=1} [\text{UCB}_{t_n+1}(\theta_{h, t_n+1},A_{h,t_n+1}) - \text{LCB}_{t_n+1}(\theta_{h, t_n+1},A_{h,t_n+1})], 
\end{align*}
and define the sigma algebra ${\cal F}_{n-1} = \sigma(\{(\theta_{h, s}, A_{h, s}, Y_{h, s})\}_{h\in [H], s\in [t_n-1]})$. It is clear that $X_n$ is ${\cal F}_n$-measurable, since $t_{n+1}-1\geq t_n+1$, knowing that there is a episode index from ${\cal J}_{T+1}$ sandwiched between $t_n, t_{n+1}$. In addition, from the previous paragraph, we have $\mathbb{E}[X_n ~|~{\cal F}_{n-1}]=0$ with certainty, since conditioned on ${\cal F}_{n-1}$ the functions $\text{UCB}_{t_n} = \text{UCB}_{t_n+1}, \text{LCB}_{t_n} = \text{LCB}_{t_n+1}$ are deterministic, and the conditional expectation is over the randomness in $((\theta_{h, t}, A_{h,t}))^H_{h=1}, ((\theta_{h, t+1}, A_{h,t+1}))^H_{h=1}$. Lastly, it is clear that $|X_n|\leq H$ with certainty. Altogether, applying the Azuma-Hoeffding inequality on $\{X_n\}^m_{n=1}$ with the filtration $\{{\cal F}_n\}^m_{n=0}$ and the fact that $m\leq \alpha T$, we see that the following inequality holds with probability $\geq 1-\delta$:
\begin{align}
    &\sum_{t\in {\cal J}_{T+1}} \sum^H_{h=1} [\text{UCB}_t(\theta_{h, t},A_{h,t}) \leq  \text{LCB}_t(\theta_{h, t},A_{h,t})] \nonumber\\
    &\leq \sum_{t\in {\cal I}_{T+1}} \sum^H_{h=1} [\text{UCB}_t(\theta_{h, t},A_{h,t}) - \text{LCB}_t(\theta_{h, t},A_{h,t})]  + H\left(1+\sqrt{\alpha T\log\frac{2}{\delta}}\right)\label{eq:final_AH1}.
\end{align}
Finally, we recall the observation that ${\cal I}_{T+1}\subset [T]\setminus {\cal J}_{T+1}$, and adding $\sum_{t\in [T]\setminus {\cal J}_{T+1}} \sum^H_{h=1} [\text{UCB}_t(\theta_{h, t},A_{h,t}) \leq  \text{LCB}_t(\theta_{h, t},A_{h,t})]$ to both sides of the inequality (\ref{eq:final_AH1}) proves the Lemma.

\section{Numeric Experiment}
We conduct synthesis numeric experiment to further validate the performance of proposed Algorithm Micmic-Opt-DP, by comparing it with algorithm SquareCBwK in \cite{HanZWXZ23} and algorithm Logistics-UCB1 (Box B) in \cite{LiS22}. Both SquareCBwK and Logistics-UCB1 are designed for a single Episode, i.e $T=1$, assuming the context distribution $\Gamma_h$ in each round index is the same. 

SquareCBwK requires an Oracle for predicting the mean reward and mean consumption for each arm, while Logistics-UCB1 requires an Oracle predicting the conversion probability and its confidence interval. To adapt them for the problem setting in this paper, always assuming $\Gamma_h$ is the same for all $h\in [H]$. 

We follow section \ref{sec:K} and \ref{sec:logistic} to set up Oracles, and follow Table \ref{tab:application} to define the function $r(\theta, a)$ and $d(\theta, a)$.
\begin{itemize}
    \item For Non-contextual Bandit, we consider First Price Auction problem in Table \ref{tab:application}.
    
    Regarding the oracle, we do not make any modification on section \ref{sec:K}. We assume the context follows uniform distribution on a discrete set $\{1, 2,\ldots,K+1\}$. The conversion probability is $\frac{a}{K+1}$ for each $a\in [K]$, independent of the context $\theta$.
    
    \item For Contextual Logistic Bandit, we consider the Pricing problem in Table \ref{tab:application}. Given the context $\theta\in [0,1]^2$ and action $a\in [K]$, the conversion probability is $\frac{1}{1+e^{-\frac{\theta_1+\theta_2-a}{\sqrt{3}}}}$. 
    
    Regarding the Oracle, we make some modification on section \ref{sec:logistic}. We take $\gamma_{N, \lambda,\delta}=0.5$ and $\kappa_f=8$, to accelerate the convergence of estimation, or the its confidence interval for the conversion probability always contains $[0, 1]$. 

    The context set $\Theta$ is $\{(n-1)  / 99\}_{n=1}^{100}\times \{(n-1)  / 99\}_{n=1}^{100}$. We assume $\theta$ follows uniform distribution in $\Theta$. 
\end{itemize}
Since in both setting, $\Theta$ only contains finite elements. We can use dynamic programming to precisely the value $\text{opt}$. Besides, we take $T=200$, $H=24$, $B=5$ in both problem instances. At the start of each episode, we reset the budget of all the algorithms as $B$. The repetition time of each both problem is 50.

At episode $t$, round $h-1$, the Oracle for algorithms SquareCBwK and Logistics-UCB1, would use the history $H =\{\theta_{h',t'}, A_{h',t'}\}_{h'=1, t'=1}^{h-1, t-1} \cup \{\theta_{h',t}, A_{h',t}\}_{h'=1}^{h-1}$ to train the model and output the predicted conversion probability or predicted mean reward. Specifically, we implement Logistics-UCB1 by taking $B_T=B$, to enhance its numeric performance.

\begin{figure}[htbp]
  \centering
  \begin{subfigure}{0.49\textwidth}
    \includegraphics[width=\linewidth]{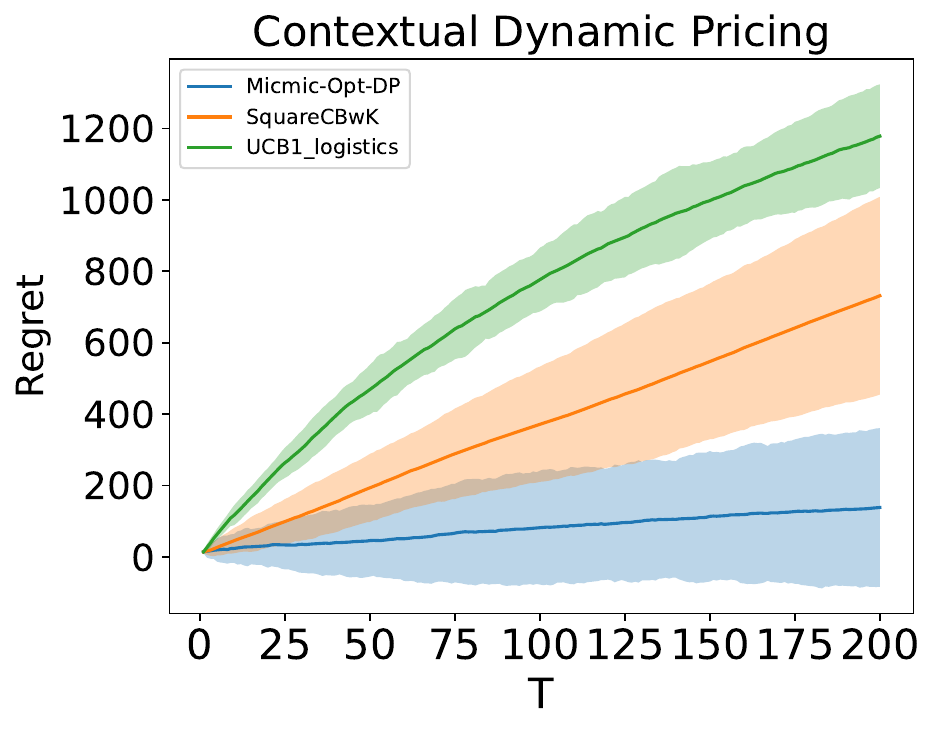}
    \label{fig:numeric_logistics}
  \end{subfigure}
  \hfill
  \begin{subfigure}{0.49\textwidth}
    \includegraphics[width=\linewidth]{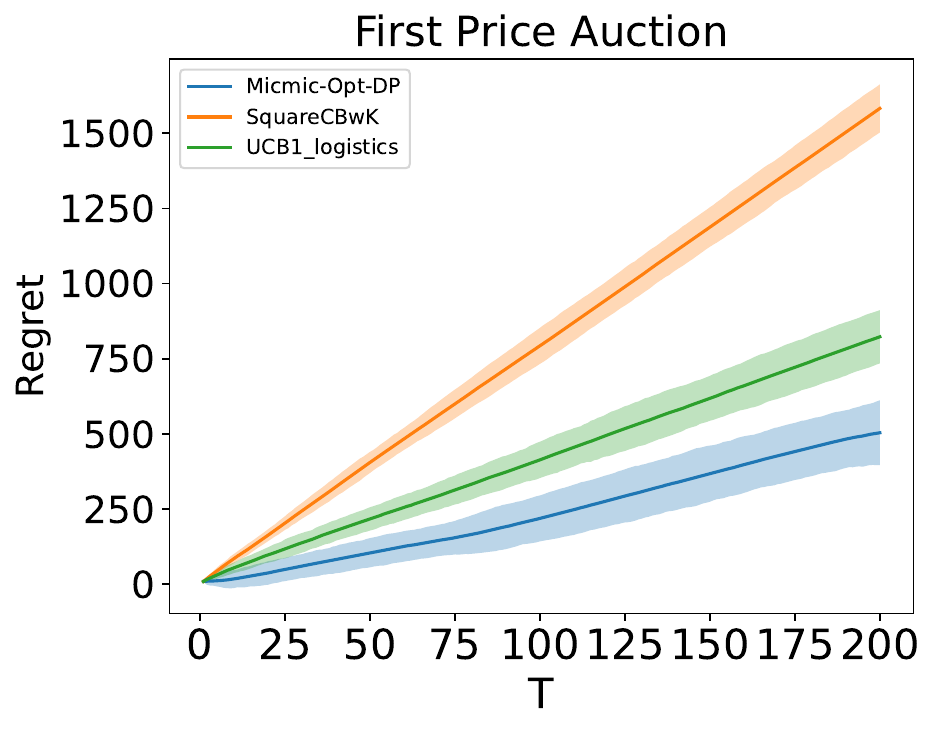}
    \label{fig:numeric_noncontextual}
  \end{subfigure}
  \caption{Numeric Performance of Micmic-Opt-DP and Benchmark Algorithms}
  \label{fig:numeric}
\end{figure}

Figure \ref{fig:numeric} presents the regret curve for each algorithm. Each point in the curve denotes the mean cumulative regret at the end of an episode. The error bar is 3 times the standard error across 50 independent experiments. 

Figure \ref{fig:numeric} presents the excellency of our proposed algorithm Micmic-Opt-DP, as the cumulative regret of the Micmic-Opt-DP is much below the others. This observation is not surprising, as the theoretical analysis for SquareCBwK and Logistics-UCB1 can only guarantee the regret $\tilde{O}(\sqrt{H})$ for $T=1$. And the regret analysis of a general $T$ remains unclear. If these two algorithm could not guarantee the convergence of optimal action in each episode, the expected additive regret in each episode must be positive, further resulting $\Omega(T)$ regret. In contrast, the theoretical analysis for the Micmic-Opt-DP asserts the sublinear regret shape regarding $T$, which is consistent with the numeric result.

\end{document}